\pdfoutput=1

\documentclass[11pt]{article}

\usepackage{ACL2023}

\usepackage{times}
\usepackage{latexsym}

\usepackage[T1]{fontenc}

\usepackage[utf8]{inputenc}

\usepackage{microtype}

\usepackage{inconsolata}

\usepackage{amssymb}
\usepackage{amsmath}
\usepackage{booktabs}
\usepackage{multicol}
\usepackage{multirow}
\usepackage{graphicx}
\usepackage{subcaption}
\usepackage{layouts}
\usepackage{enumitem}
\usepackage{makecell}
\usepackage{float}

\newcommand{\la}{$_\texttt{large}$}
\newcommand{\ba}{$_\texttt{base}$}

\title{Contextual Distortion Reveals Constituency: \\Masked Language Models are Implicit Parsers}

\author{Jiaxi Li \and Wei Lu \\
  StatNLP Research Group \\
  Singapore University of Technology and Design \\
  \texttt{jiaxi\_li@mymail.sutd.edu.sg}, \texttt{luwei@sutd.edu.sg} \\
}

\begin{document}
\maketitle
\begin{abstract}
 Recent advancements in pre-trained language models (PLMs) have demonstrated that these models possess some degree of syntactic awareness. 
 To leverage this knowledge, we propose a novel chart-based method for extracting parse trees from masked language models (LMs) without the need to train separate parsers. Our method computes a score for each span based on the distortion of contextual representations resulting from linguistic perturbations.
 We design a set of perturbations motivated by the linguistic concept of constituency tests, and use these to score each span by aggregating the distortion scores. To produce a parse tree, we use chart parsing to find the tree with the minimum score. 
 Our method consistently outperforms previous state-of-the-art methods on English with masked LMs, and also demonstrates superior performance in a multilingual setting, outperforming the state of the art in 6 out of 8 languages.
 Notably, although our method does not involve parameter updates or extensive hyperparameter search, its performance can even surpass some unsupervised parsing methods that require fine-tuning. Our analysis highlights that the distortion of contextual representation resulting from syntactic perturbation can serve as an effective indicator of constituency across languages.\footnote{Our code is available at \url{https://github.com/jxjessieli/contextual-distortion-parser}.}
\end{abstract}

\section{Introduction}

Constituency parsing is a fundamental task in natural language processing (NLP) that involves uncovering the syntactic structure of a sentence by identifying the constituents it is composed of. 
While supervised constituency parsing methods necessitate the utilization of a labeled dataset containing sentences and their corresponding constituency parses, unsupervised methods for generating syntax trees emerge because manual annotation is labor-intensive and requires specialized linguistic knowledge. 
One line of work for unsupervised constituency parsing involves designing an objective function that enables the model to infer the hierarchical structure of language from the unannotated text \citep{kim-etal-2019-unsupervised, kim-etal-2019-compound, drozdov-etal-2019-unsupervised, yang-etal-2021-pcfgs}. An alternative approach, known as Constituency Parse Extraction from Pre-trained Language Models (CPE-PLM), involves extracting parse trees from pre-trained language models without fine-tuning in an unsupervised manner \citep{Kim2020Are, wu2020perturbed, kim2021multilingual}. The main motivation for CPE-PLM is the assumption that pre-trained language models contain implicit syntactic knowledge learned during the pre-training stage. This knowledge can then be used to directly predict parse trees, eliminating the need for task-specific fine-tuning. While CPE-PLM systems have been shown to produce parse trees that resemble manually annotated ones, they have also been found to have lower performance than the first line of work. 

\begin{figure}
    \centering

    \includegraphics[width=0.72\linewidth]{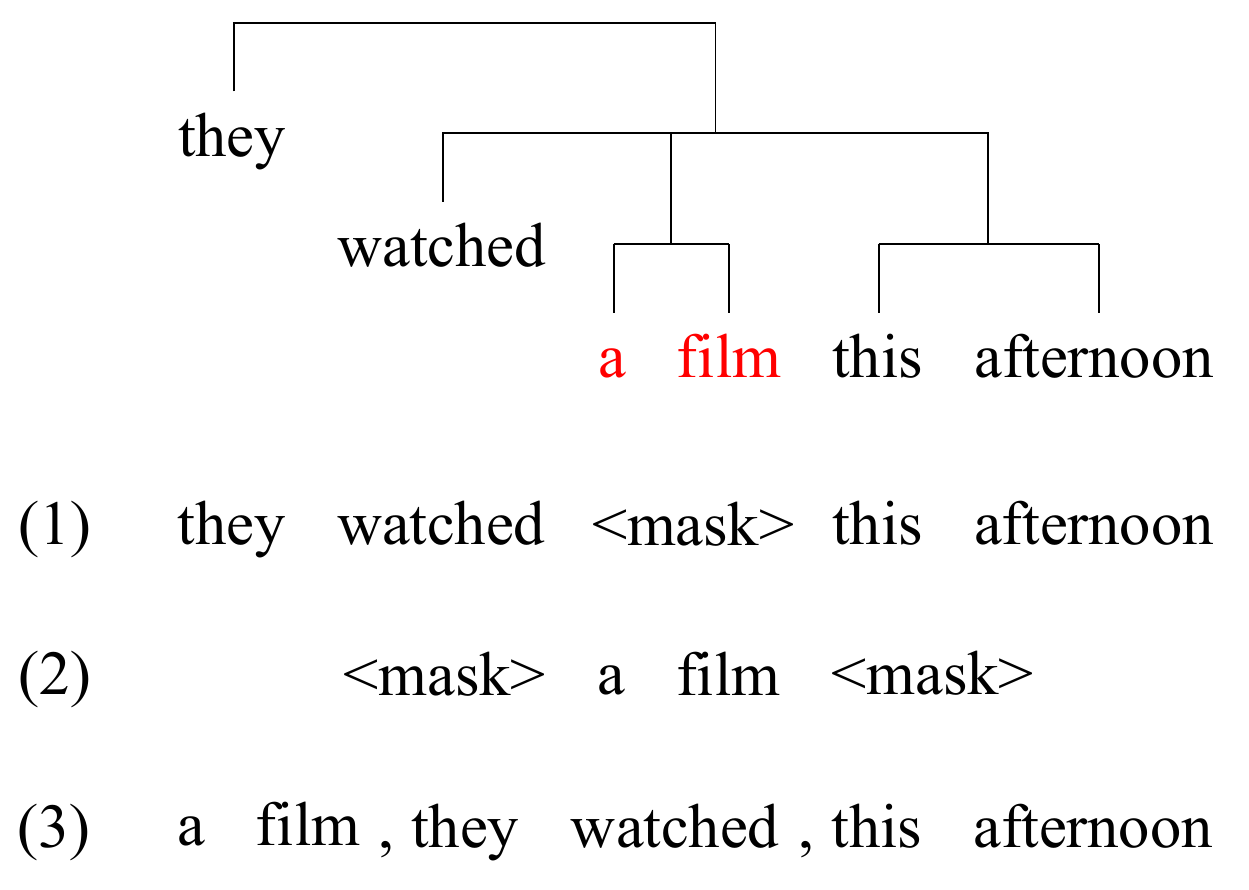}
    \caption{Example sentence and its constituency tree. We list perturbed sentences after substitution (1), decontextualization (2), and movement (3).}
    \label{fig:perturbation_example}
\end{figure}

In this paper, we propose a simple yet effective CPE-PLM approach to bridge the performance gap between these two methods by input perturbations designed based on the intuition of constituency tests. Linguists use constituency tests to determine whether a span of words forms a constituent in a sentence. One common constituency test is the substitution test which replaces the span of words with a single pronoun (such as ``\textit{it}'' or ``\textit{they}'') and checks if the sentence is still grammatical. For example, in Figure \ref{fig:perturbation_example}, the span ``\textit{a film}'' can be replaced with the pronoun ``\textit{it}'', resulting in the sentence ``\textit{they watched it this afternoon,}'' which is still grammatical. This suggests that ``\textit{a film}'' is likely a constituent. Our goal in this work is to maximally explore the capabilities of PLMs to induce grammar by themselves. Specifically, we focus on masked LMs and leverage the inherent properties of the mask token prediction pre-training objective. The main idea is to make pre-trained language models think like linguists, such that with constituency tests, span-level scores reflecting the likelihood of a span being a constituent can be obtained. 


The evaluation of constituency tests traditionally relies on grammaticality judgments. \citet{cao2020unsupervised} trained a classifier that can make grammaticality decisions with external data. In contrast, our approach assesses the degree of alternation in contextual representations resulting from manipulations akin to those used in constituency tests. 
We hypothesize that, when the context of a span is manipulated, the contextual representations of constituents will exhibit minimal alteration compared to those of distituents (non-constituents).
We refer to these manipulations as \textit{perturbations}, as our method measures the sensitivity of the representations to these changes. 
We define three perturbations and for each perturbation, we alter the input sentence and compare the representations of the perturbed sentences to that of the original. The three perturbations on an example span are illustrated in Figure \ref{fig:perturbation_example}. By applying perturbations to each span of words within the input sentence, we generate scores indicating the likelihood of each span being a constituent. 
To evaluate the effectiveness of our approach, we compare it with existing methods for extracting parse trees from PLMs without fine-tuning (Section \ref{sec:experiments}). Our model improves over the previously published best result by a large margin. In a multilingual setting, our model surpasses the previous state of the art in 6 out of 8 languages. Our model even outperforms some unsupervised parsing methods that require parameter updates, highlighting the effectiveness of our approach.

Our main contributions can be summarized as follows:
\begin{itemize}[topsep=0pt, partopsep=0pt, leftmargin=15pt, parsep=0pt, itemsep=5pt]
    \item We propose a novel, simple and effective method for extracting constituency trees from masked LMs based on linguistic perturbations.
    \item We demonstrate that our proposed method achieves new state-of-the-art results under the {\em no parameter update} setting on the standard English dataset and 6 out of 8 languages from a multilingual dataset with a significantly smaller hyperparameter search space than previous methods.
    \item Our work identifies the crucial elements that benefit the overall performance gain and highlights the potential of utilizing perturbations on masked LMs for understanding the underlying structure of language.
\end{itemize}

\section{Related Work}

\paragraph{Unsupervised Constituency Parsing.} Early works on unsupervised constituency parsing focused on building generative models such as probabilistic context-free grammars (PCFGs) \citep{carroll-charniak-1992-experiments} and constituent-context model \citep{klein-manning-2002-generative} with expectation-maximization (EM). More recent approaches have shown improvement by parameterizing PCFGs with neural networks and enhancing the model via latent variables \citep{kim-etal-2019-compound, zhu-etal-2020-return}. 
Instead of a generative model over sentences and trees, \citet{clark-2001-unsupervised} identified constituents based on span statistics. Our method is relevant to the above in that constituents appear in constituent contexts. 

Recent works have attempted to induce structural bias by constraining the flow of information in neural networks. Examples include the Parsing-Reading-Predict Network (PRPN) \citep{shen-etal-2018-neural}, the Ordered Neuron (ON) model \citep{shen-etal-2019-ordered}, and Tree transformer \citep{wang2019tree}.
Models with latent tree variables can also be seen as manipulating the information flow. The unsupervised recurrent neural network grammar (URNNG) \citep{kim-etal-2019-unsupervised} and the Deep Inside-Outside Recursive Autoencoder (DIORA) \citep{drozdov-etal-2019-unsupervised} optimized an autoencoder objective through latent tree variables.

On the other hand, \citet{cao2020unsupervised} designed an unsupervised parser by specifying a set of transformations inspired by constituency tests and trained a classifier on the external raw text of 5 million sentences from English Gigaword \citep{graff-cieri-2003-gigaword} to make grammaticality decisions. Our work builds upon this approach by utilizing constituency tests to obtain span scores. However, our method differs in that it is free from parameter updates and does not require the training of a grammaticality model on external data.

\paragraph{Constituency Parse Extraction from Pre-trained Language Models.} Inducing the parse tree of an input sentence with pre-trained language models without training is a rising line of research recently. MART \citep{wu2020perturbed} measured the impact a word has on predicting another word using BERT's hidden states and parsed by finding the best splitting point recursively. Our work is similar to their notion of perturbation and parse, while we adopt stronger prior knowledge with constituency tests and we focus on the span-level constituency. 

\citet{Kim2020Are} calculated syntactic distances of adjacent words using intermediate hidden states and the attention distributions.
\citet{li2020heads} ranked Transformer attention heads and created an ensemble of them for parsing. \citet{kim2021multilingual} and \citet{kim-2022-revisiting} further improved over \citet{Kim2020Are} by a chart-based method and top-K ensemble and extended the approach to different languages by applying multilingual PLMs. 
Our method has a clear advantage over existing approaches by leveraging the masked LMs pre-training objective implicitly with models like BERT \citep{devlin-etal-2019-bert} and RoBERTa \citep{liu2019roberta}. While previous works can be generalized to a wider range of pre-trained models, our method requires minimal hyperparameter search and consistently achieves superior results.

\section{Approach}

\subsection{Perturbations}
\label{sec:perturbations}

We specify a set of perturbations that are based on the linguistic concept of constituency tests \citep{de-marcken-1996-linguistic}.
These perturbations involve a set of transformation functions, a masked LM, and a function $d$ for calculating the distortion of the representation of a targeted span. Each transformation function takes in a sentence and a targeted span and outputs a new sentence. The masked LM takes in a sequence of words of length $T$ and outputs representations from the $l$-th layer $\mathbf{H} = \left[\mathbf{h}_1, ..., \mathbf{h}_T \right]$, where $\mathbf{h}_t \in \mathbb{R}^d$ is the contextualized representation for word $t$. The distortion function compares the change in the representation of the targeted span that results from the application of the transformation function. Our distortion function is formalized as $d: (\mathbf{H}, \tilde{\mathbf{H}}) = {||\mathbf{H} - \tilde{\mathbf{H}}||}^2 / T$, where $||\cdot||$ is the matrix 2-norm (i.e., Frobenius norm). \citet{alleman-etal-2021-syntactic} used the Frobenius norm to measure the distortion of contextual representations and observed the results of different norms to be similar. Our preliminary experiments show that the squared Frobenius norm performs slightly better than the Fronenius norm possibly due to its ability to amplify matrix differences and better distinguish constituents and distituents. We conduct an ablation study comparing these two norms in Section \ref{sec:ablation_studies}.

We focus on the sensitivity of a span's representation towards each perturbation because it may give us evidence about whether the span is a constituent. Specifically, we define three perturbations to obtain the overall distortion score for each span in a sentence. 

\paragraph{Substitution}
Substitution is a common type of constituency test that involves replacing the span of words with a single pronoun (such as ``\textit{it}'' or ``\textit{they}''). If the resulting sentence is still grammatically correct while maintaining the meaning, then the span of words forms a constituent. Instead of measuring the grammaticality change, we measure the representational distortion by substitution transformation. Specifically, the transformation function replaces
the target span of words $\left[x_i, ..., x_j\right]$ with a single mask token. Then we input the perturbed sentence 
into the pre-trained language model to obtain the representation:
\begin{equation}
    \tilde{\mathbf{H}} = \left[ \tilde{\mathbf{h}}_1, ..., \tilde{\mathbf{h}}_{i-1}, \tilde{\mathbf{h}}_{{\text{mask}}}, \tilde{\mathbf{h}}_{j+1}, ... \tilde{\mathbf{h}}_T \right].
\end{equation}

We calculate the representational distortion of the surrounding text of this span:
\begin{equation}
    d_{{sub}} = d(\mathbf{{H}} \backslash \{ \mathbf{h}_i, ..., \mathbf{h}_j \}, \tilde{\mathbf{H}} \backslash \{ \tilde{\mathbf{h}}_{{\text{mask}}} \}).
\end{equation}

The intuition is that if a span of text constitutes a grammatical unit, then the surrounding text's representation should be relatively independent of the span.
We use a single mask token to replace the targeted span as the masked language models (LMs) are pre-trained with the objective to predict the mask token. This way, while there are many choices of the single word that can replace the constituent, we allow the model to decide on the replacement word for the constituent given the context. 

\paragraph{Decontextualization}
Another way that linguists use to determine constituency is the \textit{standalone test}, also known as the \textit{answer fragment test}. It checks if the span of words can appear alone as a sentence fragment in response to a question. Our observations indicate that standalone answers often convey the same meaning as they do in their original sentence.  Therefore, we hypothesize that the representation of a constituent without context should resemble its representation in the original sentence. We decontextualize the span by masking its surrounding context instead of directly removing all the surrounding tokens. This allows us to inform the LM that the span is surrounded by an unknown context.
As the context surrounding a constituent is usually a structured unit in a sentence, we assume it can be replaced by a single word such that the meaning of the constituent changes little. Formally, for the span $\left[x_i, ..., x_j\right]$, we mask the context of it and feed it to the pre-trained model to obtain 
\begin{equation}
    \tilde{\mathbf{H}} = \left[\mathbf{h}_{\text{mask}_1}, \tilde{\mathbf{h}}_i, ..., \tilde{\mathbf{h}}_{j}, \mathbf{h}_{\text{mask}_2}\right].
\end{equation}

The distortion score is then 
\begin{equation}
    d_{dc} = d(\mathbf{H} \left[ i:j \right], \tilde{\mathbf{H}} \backslash \{ \tilde{\mathbf{h}}_{{\text{mask}_1}}, \tilde{\mathbf{h}}_{{\text{mask}_2}} \}).
\end{equation}

\paragraph{Movement}
Movement is yet another common method to determine constituency. It involves moving the span of words to a different location in the sentence and seeing if the resulting sentence is still grammatically correct. Similar to the aforementioned methods, instead of checking if the resulting sentence is grammatical, we measure the representational distortion caused by the movement transformation. We calculate distortion scores for both front movement and end movement. For a targeted span $\left[x_i, ..., x_j\right]$, the movement transformation leads to $\left[x_i, ..., x_j, x_1, ..., x_{i-1}, x_{j+1}, ..., x_{T}\right]$ and $\left[ x_{1}, ..., x_{i-1}, x_{j+1}, ..., x_{T}, x_i, ..., x_j \right]$. Then with the pre-trained language model, we obtain 
\begin{equation}
    \tilde{\mathbf{H}}_{\text{front}} = \left[\tilde{\mathbf{h}}_i, ..., \tilde{\mathbf{h}}_j, \tilde{\mathbf{h}}_1, ..., \tilde{\mathbf{h}}_{i-1}, \tilde{\mathbf{h}}_{j+1}, \tilde{\mathbf{h}}_{T}\right]
\end{equation}
\begin{equation}
    {\mathbf{H}^{\prime}}_{\text{end}} = \left[\mathbf{h}^\prime_1, ..., \mathbf{h}^\prime_{i-1}, \mathbf{h}^\prime_{j+1}, ..., \mathbf{h}^\prime_{T}, \mathbf{h}^\prime_{i}, ..., \mathbf{h}^\prime_{j}\right]
\end{equation}

Each movement splits the sentence into three spans. To make the split more explicit to the pre-trained language model, we add a comma between spans to separate them. We calculate the distortion of each movement of the span by summing up the three distortion scores. Therefore, the distortion score is 
\begin{equation}
    \begin{split}
    d_{move} & = d_{frontmove} + d_{endmove} \\
    & = d(\mathbf{H}\left[ 1:i-1 \right], \tilde{\mathbf{H}}\left[ 1:i-1 \right]) \\
    & + d(\mathbf{H}\left[ i:j \right], \tilde{\mathbf{H}}\left[ i:j \right]) \\
    & + d(\mathbf{H}\left[ j+1:T \right], \tilde{\mathbf{H}}\left[ j+1:T \right]) \\
    & + d(\mathbf{H}\left[ 1:i-1 \right], \mathbf{H}^{\prime}\left[ 1:i-1 \right]) \\
    & + d(\mathbf{H}\left[ i:j \right], \mathbf{H}^{\prime}\left[ i:j \right]) \\
    & + d(\mathbf{H}\left[ j+1:T \right], \mathbf{H}^{\prime}\left[ j+1:T \right]).
\end{split}
\end{equation}


For each span in a sentence, we apply the aforementioned three perturbations and score each span by averaging the span-level contextual distortion yielded from perturbations\textcolor{red}{\footnote{Sequentially calculating the attention matrix for a sentence of length $n$ is $O(n^4)$ computational complexity, but GPUs' parallel processing capabilities allow for $O(1)$ sequential operations per tested span (as detailed in Section $4$ of \citet{NIPS2017_3f5ee243}). Therefore, the time complexity to obtain the distortion scores for a sentence of length $n$ with parallel attention matrix computation is $O(n^2)$.}} 
\begin{equation}
    d = \frac{1}{L} (d_{sub} + d_{dc} + d_{move}),
\end{equation}
where $L$ denotes the number of span-level contextual distortion scores.\footnote{When the span is in the middle of the sentence, the three perturbations produce 8 span-level scores, with the movement perturbation contributing 6 of these scores. In contrast, when spans are not in the middle of the sentence, the movement perturbation yields only 4 scores as each front and end move splits the span by a comma into two. Therefore, averaging the scores mitigate this bias caused by the position of the spans.}
It is worth noting that the decontextualization and movement perturbations align with the intuition of \citet{wu2020perturbed}. 
They suggest that words within a constituent have significant interaction with each other, while words that are syntactically far apart have minimal interaction.
In our approach, we assume that the representation of a word is primarily affected by its syntactically local context, i.e., the constituent that it appears in. 

\subsection{Parsing Algorithm}\label{sec:parser}

With the distortion score calculated for each span in a sentence, we describe how to obtain the parse tree in this section. In the supervised setting, \citet{stern-etal-2017-minimal} and \citet{kitaev-klein-2018-constituency} showed that independently scoring each span and then choosing the tree with the best total score produced a simple yet very accurate parser. We apply a similar chart parsing approach to obtain the best tree given the span scores. 

We define the score $s(T)$ of a tree $T$ to be the sum of its normalized distortion scores denoted as $\hat{d}(i, j)$ spanning words $i$ to $j$,  
\begin{equation}
    s(T) = \sum_{(i, j) \in T} \hat{d}(i, j).
\end{equation}

One important step to make chart parsing work with our distortion scores is normalization. Our perturbations invoke inevitable bias towards the length of the span. If the length of the target span is relatively long, the distortion score will generally be small compared to shorter spans regardless of the constituency of the span, while distortion scores of spans of the same length are comparable with each other. Therefore, we normalize the distortion score over span lengths in each sentence such that scores of the same length are scaled individually to the unit norm. For span $(i, j)$, whose distortion score is $d(i, j)$, the normalized distortion score is:
\begin{equation}
\label{eq:normalization}
    \hat{d}(i, j) = \frac{d(i, j)}{\sqrt{\sum\limits_{{(i',j')} \text{s.t.} {j'-i'=j-i}} d^2(i', j')}}
\end{equation}

Since the distortion score is inversely proportional to the likelihood of a span being a constituent, we need to find the minimum-scoring tree. As with chart parsing with the standard CKY algorithm, the running time of this procedure is $O(n^3)$ for a sentence of length $n$.


The best score of a tree spanning $i$ to $j$ with $k$ as the splitting point is defined to be the sum of the scores of the two subtrees and the current span's normalized distortion score.
The recurrence relation used for finding the tree with the best score $s^{*}$ spanning $i$ to $j$ is:
\begin{equation}
    s^{*}(i, j) = \min_{k} \left[ s^{*} (i, k) + s^{*} (k, j) + \hat{d}(i, j) \right]
\end{equation}

Based on these optimal scores, we will then be able to use a top-down backtracking process to arrive at the optimal constituency tree, as our output.

\section{Experiments}
\label{sec:experiments}
\subsection{Setup}

We conduct experiments on the English Penn Treebank (PTB) dataset \citep{marcus-etal-1993-building}.
To understand how our approach works across different languages, following prior research \citep{kim2021multilingual, zhao-titov-2021-empirical}, we also evaluate our approach on 8 different other languages, namely Basque, French, German, Hebrew, Hungarian, Korean, Polish, and Swedish, which are freely released within the SPMRL dataset\footnote{Dataset statistics can be found in Appendix \ref{appendix:dataset_stats}.} \citep{seddah-etal-2013-overview}.
The evaluation was performed using the F1 score, which was calculated with respect to the gold trees in the PTB test set (section 23) and the test sets for different languages in SPMRL\footnote{Following prior research \citep{kim-etal-2019-compound, Kim2020Are}, punctuation was removed, unary chains were collapsed before evaluation, the F1 score was calculated disregarding trivial spans, and the results reported are based on the unlabeled sentence-level F1.}.

\subsection{Implementation Details}

For the English PTB dataset, the results of masked language models (BERT and RoBERTa) were reported. For the multilingual SPMRL dataset, the results of a multilingual version of the BERT-base model (M-BERT \citet{devlin-etal-2019-bert})\footnote{We use \texttt{bert-base-multilingual-uncased}.} were reported\footnote{More details can be found in Appendix \ref{sec:addtional_implementation_details}.}. Our method uses a single hyperparameter, the layer of representation,
and we select the optimal layer for each LM by evaluating parsing performance on the development set.  

\begin{table*}[t]
    \begin{center}
    \centering
    \resizebox{0.9\linewidth}{!}{
    \small
    \begin{tabular}{llcccccccc}
    \toprule
        \textbf{Model} & \textbf{Method}  &\textbf{Layer}  & \textbf{S-F1}   & \textbf{SBAR}  &\textbf{NP}    & \textbf{VP}    & \textbf{PP}    & \textbf{ADJP}  & \textbf{ADVP}  \\
    \midrule
    \multirow{2}{*}{Baselines} &
        Right-Branching                 & -     & 39.8  & 69    & 25    & 72    & 42    & 28    & 38 \\
        & Left-Branching                & -     & \textcolor{white}{0}9.0   & \textcolor{white}{0}6     & 11    & \textcolor{white}{0}1     & \textcolor{white}{0}5     & \textcolor{white}{0}3     & \textcolor{white}{0}8 \\
    \midrule
    \multirow{6}{*}{BERT\ba} &
        \citet{Kim2020Are} (w/o bias)     & 9     & 32.4      & 28    & 42    & 28    & 31    &35     & 63 \\
        & \citet{Kim2020Are}     & 9     & 42.3      & 45    & 46    & 49    & 43    & 41    & 65 \\ 
        & MART \citep{wu2020perturbed}  & 12    & 42.1      & 52    & 45    & 47    & 51    & 48    & 57  \\
        & \citet{kim2021multilingual}   & -     & 42.7      & -     & -     & -     & -     & -     & - \\
        & \citet{kim-2022-revisiting}   & -     & 43.0      & -     & -     & -     & -     & -     & - \\
        & Ours                          & 10    & \textbf{49.0}      & 49   & 65    & 39    & 74    & 46  & 64 \\ 
    \midrule
    \multirow{6}{*}{BERT\la} &
        \citet{Kim2020Are} (w/o bias)     & 17    & 34.2    & 34    & 43    & 27    & 39    & 37    & 57 \\
        & \citet{Kim2020Are}     & 17    & 44.4    & 55    & 48    & 48    & 52    & 41    & 62 \\
        & MART \citep{wu2020perturbed}  & 16    & 42.9    & 50    & 47    & 49    & 50    & 46    & 57  \\
        & \citet{kim2021multilingual}   & -     & 44.6    & -     & -     & -     & -     & -     & - \\
        & \citet{kim-2022-revisiting}   & -     & 45.0    & -     & -     & -     & -     & -     & - \\
        & Ours                          & 15    & \textbf{48.2}  & 50   & 62    & 42    & 69    & 46    & 64 \\
    \midrule
    \multirow{6}{*}{RoBERTa\ba} &
        \citet{Kim2020Are} (w/o bias)     & 9     & 33.8  & 40    & 38    & 33    & 43    & 42    & 57 \\
        & \citet{Kim2020Are}     & 8     & 42.1  & 51    & 44    & 44    & 55    & 40    & 66 \\
        & MART \citep{wu2020perturbed}  & 12    & 42.2  & 52    & 44    & 50    & 51  & 46     & 56  \\
        & \citet{kim2021multilingual}   & -     & 45.0  & -     & -     & -     & -     & -     & - \\
        & \citet{kim-2022-revisiting}   & -     & 45.4  & -     & -     & -     & -     & -     & - \\
        & Ours                          & 11    & \textbf{46.7}  & 52   & 58    & 41    & 63    & 47    & 58 \\ 
    \midrule
    \multirow{6}{*}{RoBERTa\la} &
        \citet{Kim2020Are} (w/o bias)     & 14    & 34.1  & 29    & 46    & 30    & 37    & 28    & 40 \\
        & \citet{Kim2020Are}     & 12    & 42.3  & 40    & 50    & 43    & 44    & 48    & 56 \\
        & MART  \citep{wu2020perturbed} & 24    & 41.3  & 49    & 43    & 47    & 50    & 44     & 58  \\
        & \citet{kim2021multilingual}   & -     & 42.8  & -     & -     & -     & -     & -     & - \\ 
        & \citet{kim-2022-revisiting}   & -     & 47.2  & -     & -     & -     & -     & -     & - \\
        & Ours                          & 21    & \textbf{48.8}  & 55   & 61    & 43    & 71    & 49    & 59 \\ 
    \midrule
    \multirow{5}{*}{\makecell[l]{Other models with \\ parameter update}} &
        PRPN (tuned) \citep{shen-etal-2018-neural}     & -     & 47.3  & 50    & 59    & 47    & 57    & 44    & 33 \\
        & ON (tuned) \citep{shen-etal-2019-ordered}    & -     & 48.1  & 51    & 65    & 41    & 54    & 38    & 32 \\
        & N-PCFG \citep{kim-etal-2019-compound}        & -     & 50.8  & 53    & 71    & 34    & 59    & 33    & 46 \\
        & C-PCFG \citep{kim-etal-2019-compound}        & -     & 55.2  & 56    & 75    & 42    & 69    & 40    & 53 \\
        & CT (w/o self-training) \citep{cao2020unsupervised} & -    & 48.2  & 23    & 60    & 33    & 57    & 66    & 62 \\
        
    \bottomrule
    \end{tabular}
    }
    \caption{
    \label{tab:main-performance}
    Parsing performance (S-F1) and label recall on English PTB with four masked LMs. We present results of \citet{Kim2020Are} with and without the right-branching bias. BERT{\ba} results for the MART method are from \citet{wu2020perturbed}. We run their code to produce results for other models by changing the masked LM. \citet{kim-2022-revisiting} proposed 5 variations to their method and we present the one with the best-averaged performance across LMs.
    }
    \end{center}
\end{table*}

\subsection{Parsing Performance on PTB}
Table \ref{tab:main-performance} presents the F1 scores obtained by our method in comparison to existing parsers that use pre-trained masked language models without undergoing parameter updates. It can be observed that our method consistently achieves superior performance compared to state-of-the-art methods under the same condition, with a substantial margin of improvement. It is noteworthy that our method has a significantly reduced search space for hyperparameters, with the layer index being our only hyperparameter. This is in contrast to the approach proposed by \citet{Kim2020Are},  \citet{kim2021multilingual}, and \citet{kim-2022-revisiting} which have a larger number of hyperparameters to optimize including attention heads, layer, and distance metric, etc. 

\citet{wu2020perturbed} has the same hyperparameter search space as ours. They use a top-down approach to find the split point iteratively based on the ``impact matrix'', which captures the impact of inter-word relationships. Our method focuses on the span-level information and thus might be more suitable for constituency parsing, whereas their method may be more suitable for dependency parsing, because the constituency tree is more concerned with the syntactic role of words, by grouping them into constituent spans, while the dependency tree is more concerned with the grammatical relationship between words, by connecting them with edges.

Additionally, our method even surpasses some unsupervised constituency parsing methods with parameter-update including PRPN \citep{shen-etal-2018-neural} and ON \citep{shen-etal-2019-ordered}. Our best result is approaching Neural PCFG (N-PCFG) and Compound PCFG (C-PCFG) \citep{kim-etal-2019-compound}. Notably, our best model even outperforms \citet{cao2020unsupervised} without self-training, where they used an external unlabeled large dataset to train a grammar model built on top of RoBERTa-base with additional parameters.\footnote{When iteratively refining their model using the self-training strategy involving multiple iterations of parameter updates, they achieved an average S-F1 of 62.8. While our method is not comparable to such an approach, we believe our results can also be further boosted using a similar strategy, especially when a parameter-rich model (e.g., RNNG) is used for this step. We focus on establishing a strong parameter-update-free approach in this work and leave such a direction to future explorations.}

In addition to sentence-level F1 (S-F1), we report label recall scores for six main types, namely SBAR, NP, VP, PP, ADJP, and ADVP. Notice that our model can significantly outperform other models in terms of the label recall of NP, PP, and ADVP as the semantic meaning of these phrases is usually independent of the context and our method is able to capture their representational change. While for other constituent types, we obtain comparable label recalls. This demonstrates that our method is effective in recognizing the main constituency types. In Section \ref{sec:analysis_perturbations}, we conduct a more detailed analysis of perturbations that lead to improvements in performance for different constituency types.

\subsection{Parsing Performance on SPMRL}
\begin{table*}[t]
    \begin{center}
        \centering
        \resizebox{0.9\linewidth}{!}{
        \small
        \begin{tabular}{lcccccccc|c}
            \toprule
                \textbf{Method}  & \textbf{Basque}    & \textbf{French}    & \textbf{German}    & \textbf{Hebrew}    & \textbf{Hungarian} & \textbf{Korean}    & \textbf{Polish}    & \textbf{Swedish}  & \textbf{Average}   \\
            \midrule
                N-PCFG$^{\ddagger}$  & 30.2      & 42.2      & 37.8      & 41.0      & 37.9      & 25.7      & 31.7           & 14.5    & 32.6      \\
                C-PCFG$^{\ddagger}$  & 27.9      & 40.5      & 37.3      & 39.2      & 38.3      & 27.7      & 32.4      & 23.7     & 33.4      \\
                \citet{kim2021multilingual}     & 41.6      & 45.6      & 40.3      & 42.0      & \textbf{40.4}      & \textbf{49.8}      & 42.9      & 39.3     & 42.7      \\
                \citet{kim-2022-revisiting}     & 41.2  & 36.1  & 37.6  & 38.0  & 33.8  & 49.1  & 51.4  & 32.6 & 40.0 \\
                Ours    & \textbf{44.0}      & \textbf{48.7}      & \textbf{40.8}      & \textbf{50.4}      & 39.1      & 43.7      & \textbf{53.3}      & \textbf{46.3}    & \textbf{45.8}      \\
            \bottomrule
        \end{tabular}
        }
        \caption{
            \label{tab:spmrl_performance}
            Sentence F1 on the test set of 8 languages from the SPMRL dataset. ${\ddagger}$: results from \citet{zhao-titov-2021-empirical}.
        }
        
    \end{center}
\end{table*}

Table \ref{tab:spmrl_performance} shows the F1 scores on 8 languages from the SPMRL dataset. 
As baselines, we consider N-PCFG and C-PCFG, following prior work \citep{kim2021multilingual} as they can subsume naive baselines such as right or left-branching\footnote{We report the results from \citet{zhao-titov-2021-empirical} where they assume they have access to parses in the PTB development set to select the best hyperparameters following \citet{kim-etal-2019-compound}. This setting is the same as ours.}. 
From the table, it can be observed that our method outperforms the previous state-of-the-art under the setting of no parameter update in 6 out of 8 languages and with significantly fewer hyperparameters\footnote{In our case we only have one hyperparameter, while \citet{kim-2022-revisiting} and \citet{kim2021multilingual} find the best combinations of attention head, layer, distance metric and create an ensemble of attention heads. We present the results from their work where the same pre-trained LM, M-BERT is used. They proposed multiple variations of their method and we present the one with the best-averaged performance. Note that \citet{kim2021multilingual} proposed an ensemble of multiple PLMs to obtain better results and are not comparable with ours. }. Specifically, for Hebrew, Polish and Swedish, our method improves over the previous state-of-the-art by a large margin. Our average performance on 8 languages achieves 45.8 F1, an absolute improvement of 3.1 points over the previous best-published results under the same condition. Results of all languages surpass those of N-PCFG by 13.2 points and C-PCFG by 12.4 points on average. Overall, these results show that our method is robust and effective as compared to previous approaches across languages.

\subsection{Ablation Studies}
\label{sec:ablation_studies}
\paragraph{Operations to Combine Scores}
Distortion scores for each span are computed by summing scores from three distinct perturbations, an approach inspired by \citet{cao2020unsupervised}. This summation operation assumes that different perturbations may provide complementary information, which could capture a wider range of constituency evidence. Aside from summation, alternative strategies to combine the scores generated by perturbations could be the minimum (assuming a span is a constituent if one test is decisive) or the maximum (assuming a span is not a constituent if at least one test is inconclusive). We conducted further experiments using the minimum and maximum methods to combine perturbation scores. As the scores from different perturbations may be on different scales, we normalized the scores produced by each perturbation before combining them. As shown in Tables \ref{tab:ablation_min_max_fnorm_ptb} and \ref{tab:ablation_min_max_fnorm_spmrl}, these alternative methods perform less effectively than the summation method. One possible explanation is that the minimum or maximum methods may be overly sensitive to individual perturbations, potentially leading to an underestimation or overestimation of the true constituency score. In contrast, the summation method captures a more comprehensive view of the perturbations, thus reducing the influence of any single perturbation and enhancing overall scoring robustness. Our experiments also showed that whether the scores were normalized after or before the summation process did not significantly affect the results on the PTB dataset, and the latter even slightly improved the results for German, Hungarian, and Korean languages in the SPMRL dataset. These findings suggest that the method of summation, in calculating contextual distortion scores, serves as a robust mechanism for discerning constituents and distituents and that all perturbations contribute to effectively determining a span's constituency.

\begin{table}[t!]
    \centering
    \resizebox{0.93\linewidth}{!}{
    \small
    \begin{tabular}{lcccc|c}
    \toprule
        \textbf{Model} & \textbf{sum+N} & \textbf{N+sum} & \textbf{N+min} & \textbf{N+max} & \textbf{F-norm}   \\
    \midrule
        BERT\ba     & 49.0        & 48.9  & 41.6  & 40.7    & 47.6              \\
        BERT\la     & 48.2        & 48.9  & 43.1  & 41.7    & 46.5              \\
        RoBERTa\ba  & 46.7        & 46.5  & 37.0  & 41.3    & 46.3              \\
        RoBERTa\la  & 48.8        & 46.3  & 38.6  & 39.7    & 47.5              \\
    \bottomrule
    \end{tabular}
    }
    \caption{
        \label{tab:ablation_min_max_fnorm_ptb}
        Parsing performance with perturbation combinations operations and F-norm as the matrix norm on the PTB test set. sum+N indicates summing the scores first before normalization, while N+sum means normalizing scores from each perturbation and then applying the sum operation.
    }
\end{table}

\begin{table}[t!]
    \centering
    \resizebox{0.93\linewidth}{!}{
    \small
    \begin{tabular}{lcccc|c}
    \toprule
        \textbf{Language} & \textbf{sum+N} & \textbf{N+sum} & \textbf{N+min} & \textbf{N+max} & \textbf{F-norm}   \\
    \midrule
        Basque      & 44.0        & 41.9  & 36.8  & 36.6    & 45.1              \\
        French      & 48.7        & 48.7  & 35.9  & 42.6    & 50.3              \\
        German      & 40.8        & 45.8  & 35.0  & 42.4    & 40.3            \\
        Hebrew      & 50.4        & 50.4  & 37.3  & 43.2    & 51.1              \\
        Hungarian   & 39.1        & 41.9  & 30.6  & 37.9    & 38.1             \\
        Korean      & 43.7        & 44.8  & 40.6  & 40.3    & 41.6            \\
        Polish      & 53.3        & 52.0  & 44.9  & 45.2    & 53.8              \\
        Swedish     & 46.3        & 45.8  & 36.7  & 39.8    & 45.8             \\
    \midrule
        Average     & 45.8        & 46.4  & 37.2  & 41.0    & 45.8            \\
    \bottomrule
    \end{tabular}
    }
    \caption{
        \label{tab:ablation_min_max_fnorm_spmrl}
        Parsing performance with perturbation combinations operations and F-norm as the matrix norm on the SPMRL test set. 
    }
\end{table}

\paragraph{Matrix Norms}
We present a comparison between the Frobenius norm and its squared variant for distortion calculation in contextual representations, motivated by superior preliminary results from the squared variant on the PTB dataset. Tables \ref{tab:ablation_min_max_fnorm_ptb} and \ref{tab:ablation_min_max_fnorm_spmrl} show that the squared Frobenius norm consistently surpasses the conventional one on the PTB test set, though its efficacy on the SPMRL dataset is more varied. Despite this discrepancy, the inherent property of the squared Frobenius norm -- its capacity to amplify the divergence between matrices -- could potentially enable more precise identification of subtle yet important distinctions, such as those between constituents and distituents.

\section{Analysis}
\subsection{Performance Comparison by Layer}

\begin{figure}[t]
  \centering
  \begin{minipage}[c]{1.0\linewidth}
    \centering
    \begin{subfigure}{0.45\textwidth}
      \centering
      \includegraphics[width=\textwidth]{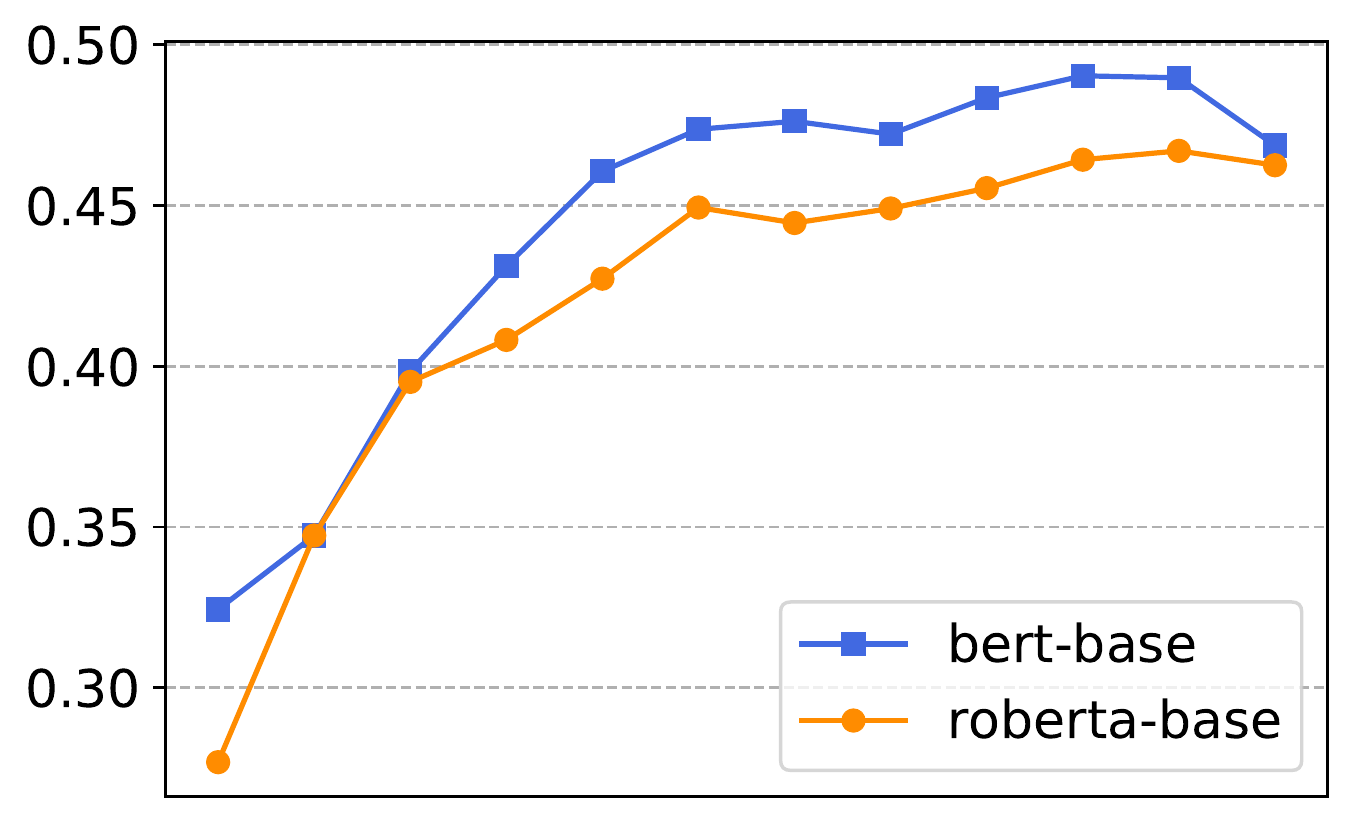}
      \captionsetup{font={scriptsize}}
      \caption{\label{fig:base_f1_layers} Base models with 12 layers. }
    \end{subfigure}
    \begin{subfigure}{0.45\textwidth}
      \centering
      \includegraphics[width=\textwidth]{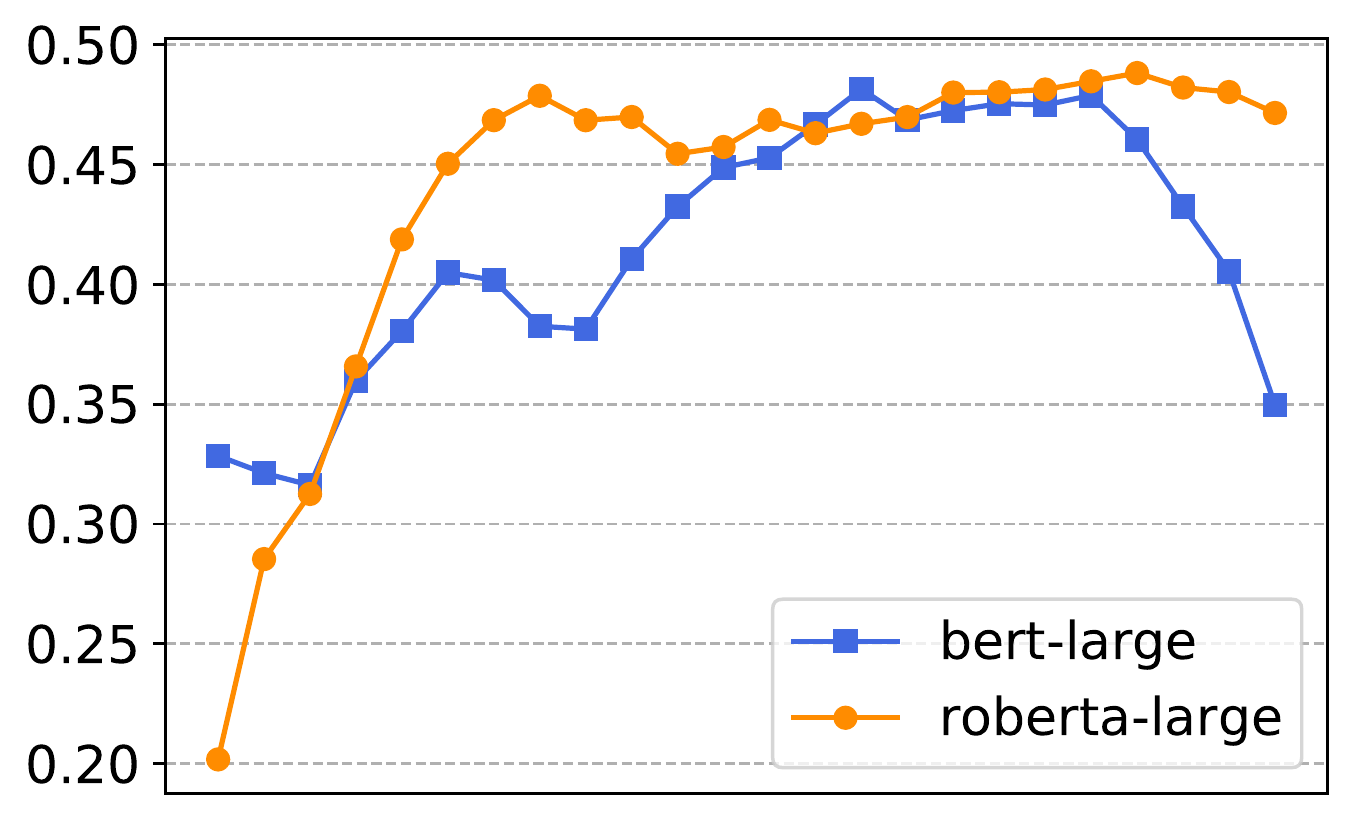}
      \captionsetup{font={scriptsize}}
      \caption{\label{fig:large_f1_layers} Large models with 24 layers. }
    \end{subfigure}
    \caption{The layer-wise sentence-level F1 scores on the PTB test set.}
    \label{fig:f1_layers_ptb}
  \end{minipage}\hfill
\end{figure}

To gain a better understanding of the relationship between the layers of pre-trained LMs and parsing performance, we plot the layer-wise F1 scores on the PTB test set in Figure \ref{fig:f1_layers_ptb}.
We can observe several patterns in the figure. First, the best-performing layers are largely found in the later layers of the LMs, but not necessarily the last layer. In our approach, the representation from LMs is mainly used for the semantic information it contains as we focus on the change of their contextual meaning, and it is likely that the later layers contain richer semantic information.
However, we notice there is a performance drop for BERT-large when later layers are considered. 
We suspect this is because BERT-large is more prone to be undertrained compared to BERT-base and the deeper layers may contain more noise. The issue is not present in RoBERTa models probably because they are pre-trained on a much larger dataset with carefully designed pre-training strategies. 
Second, we note that the best-performing layers on the development set for different languages are relatively consistent\footnote{With the same multilingual model, the best-performing layers on 8 languages from SPMRL are usually 10 or 11. Further details of performance comparison by layer on the SPMRL dataset can be found in Appendix \ref{appendix:performance_layer}.}.  This suggests that there are specific layers, typically the later layers in the LMs, which are more sensitive to linguistic perturbations and can reflect the information of constituency with our perturbations. 

\subsection{Impact of Perturbation Types}
\label{sec:analysis_perturbations}

\begin{figure*}[t]
    \centering
    \includegraphics[width=0.75\linewidth]{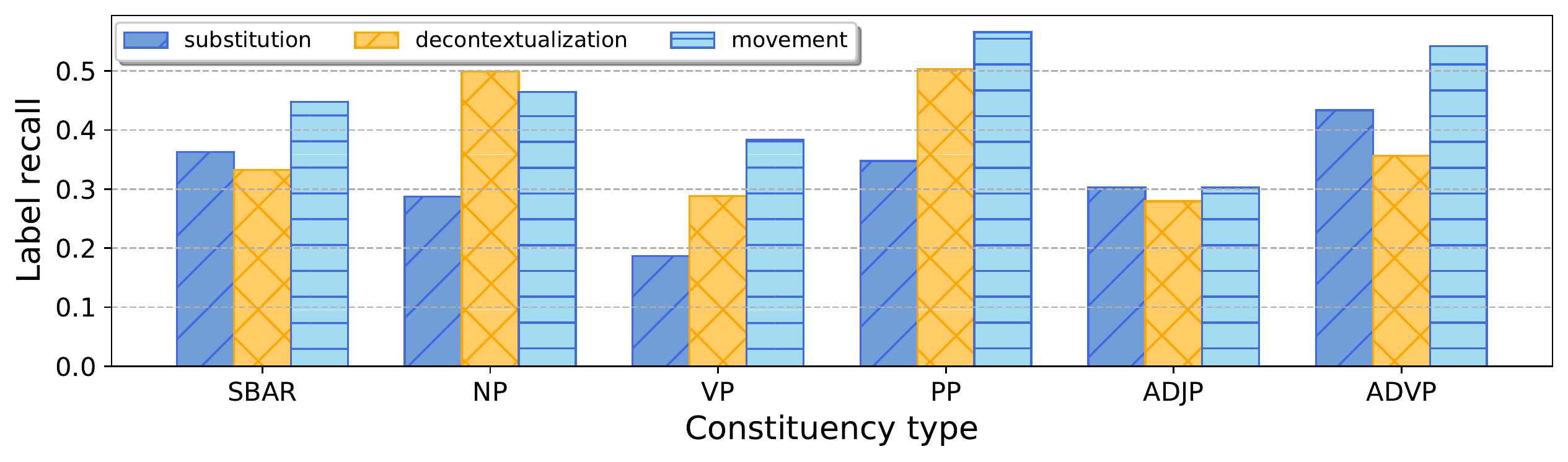}
    \caption{Label recall scores of 6 main constituency types when one perturbation is applied at a time. We use the BERT-base model on the English PTB test set as an illustration.}
    \label{fig:label_recall_bert_base}
\end{figure*}

\begin{table}[t!]
    \centering
    \resizebox{0.9\linewidth}{!}{
    \small
    \begin{tabular}{lcccc}
    \toprule
        \textbf{Model} & \textbf{w/o sub} & \textbf{w/o dc} & \textbf{w/o move} & \textbf{all}   \\
    \midrule
        BERT\ba     & 47.2      & 44.2      & 40.9  & \textbf{49.0}              \\
        BERT\la     & 45.3      & 42.0      & 45.4  & \textbf{48.2}              \\
        RoBERTa\ba  & 43.9      & 44.4      & 41.8  & \textbf{46.7}              \\
        RoBERTa\la  & 46.9      & 46.8      & 40.2  & \textbf{48.8}              \\
    \bottomrule
    \end{tabular}
    }
    \caption{
        \label{tab:ablation_perturb}
        Parsing performance with perturbation combinations on the PTB test set. 
    }
\end{table}

Our method aggregates three types of perturbations to obtain improved results. In this section, we analyze the impact of perturbation combinations. Specifically, we first conduct an ablation study to verify that each perturbation helps to improve the overall parsing performance. Then for each perturbation, we examine the constituency types that it can extract effectively. Table \ref{tab:ablation_perturb} illustrates the sentence-level F1 score on the PTB test set when removing one type of perturbation each time\footnote{Results on other languages can be found in Appendix \ref{appendix:impact_perturbation_types}}. For each language model, the layer that generates the best parsing results on the development set is used.
From the results presented in Table \ref{tab:ablation_perturb}, we observe that each perturbation contributes to the improvement of the results. Additionally, each PLM has a different sensitivity to different perturbations. For example, the performance drops the most when the movement perturbation is not used, except for BERT-large. We find that although BERT-large is effective in extracting constituency trees with our method, the patterns such as layer-wise performance and perturbation combination performances are different from those of other models. We believe that the differences in representation between BERT-large and other masked LMs could be an interesting research question worth exploring further.

Figure \ref{fig:label_recall_bert_base} illustrates the label recall of 6 main constituency types when one perturbation is applied at a time\footnote{We conduct our analysis on the BERT-base model. The full results can be seen in Appendix \ref{appendix:impact_perturbation_types}.}. It can be observed that the movement perturbation is generally more effective in capturing all types of constituents compared to the other perturbations. We note that each constituent type can be effectively captured by at least one perturbation and each perturbation targets different constituency types. For example, with movement perturbation, SBAR, NP, PP, and ADVP have high label recalls. This is likely because when the position of these constituents is changed within a sentence, the meaning of the phrase itself, and the context around it that are reflected in the span representations normally remain unchanged. For NP and PP, decontextualization is effective because the contextual representations of these phrases are primarily determined by themselves. With or without context has a relatively small impact on the contextual representations of these constituents. Substitution works well for SBAR, PP, and ADVP, as these phrases can usually be replaced by a single word without causing the meaning of the surrounding context to be altered.

\subsection{Distortion Score Reveals Constituency}

\begin{figure}[t]
    \centering
    \includegraphics[width=0.8\linewidth]{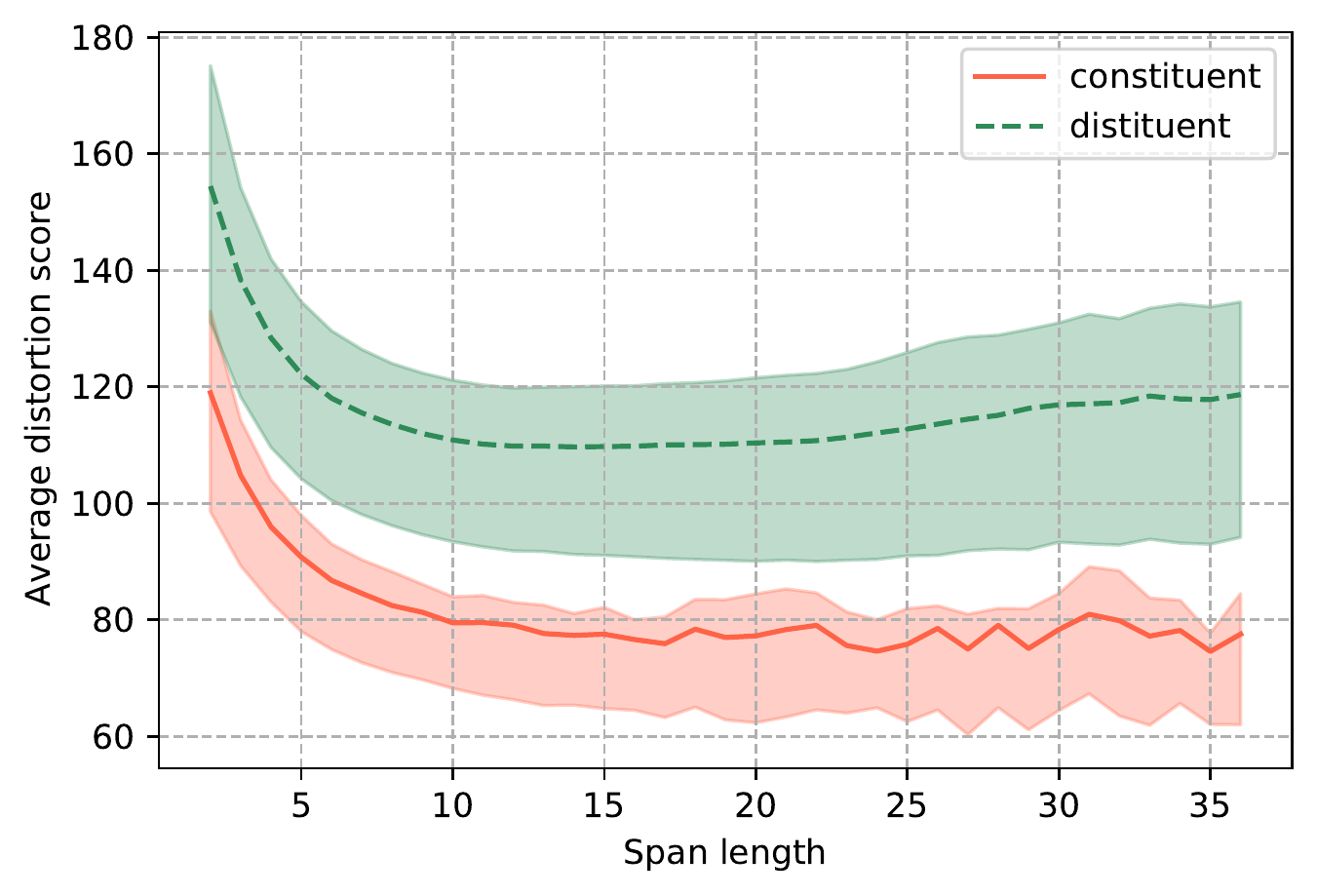}
    \caption{Distortion scores of constituents and distituents with the BERT-base model. }
    \label{fig:distortion_length_bert_base}
\end{figure}

We further analyze the correlation between distortion score and constituency.
We collect the distortion scores before normalization for each constituent and distituent in the gold sentences in the test set of PTB. Figure \ref{fig:distortion_length_bert_base} illustrates the distortion scores for constituents and distituents of varying span lengths\footnote{To ensure that there is enough data for analysis, we restrict our analysis to spans whose lengths are less than or equal to 36, as PTB does not have enough longer constituents.}. 
 We group the spans by their lengths, and for each group of spans, the shaded areas represent the 30th to 70th percentile range of distortion scores for each group.
It can be seen that the distortion scores for constituents are normally smaller than those of distituents, which verifies our hypothesis that distortion scores of representations calculated with perturbation reveal the likelihood of a span being a constituent. 

As observed in Figure \ref{fig:distortion_length_bert_base}, distortion scores for spans of the same lengths are comparable to each other, but not so when the spans are of different lengths. The distortion score indicates the likelihood that one span is a constituent compared to the other when both spans have the same length. However, when the two spans are of different lengths, the longer span is likely to have a lower distortion score, not because it is a constituent, but due to the lesser amount of perturbed information. It is therefore essential to apply normalization over the length of spans as shown in Equation \ref{eq:normalization}.

\section{Conclusion}

In this work, we proposed a novel method for extracting constituency trees from masked LMs without parameter updates. Based on linguistic perturbations, we use the change in the contextual representation to reveal the constituency property of a span.
Through experiments on the English PTB and the multilingual SPMRL dataset, we show that our method is robust and able to achieve state-of-the-art performance across languages.
Notably, our method only requires a single hyperparameter, the layer index within the Transformer architecture. Our results indicate that our method is a simple yet effective approach to obtaining constituency trees, and future research includes exploring its application to broader PLMs beyond masked LMs and the identification of other types of syntactic structures. 

\section*{Limitations}
Our method has some limitations that should be acknowledged and addressed in future research. One of the main limitations is the restriction of the PLM type to masked LMs. While this model type has been widely used in previous studies, it may not be the only option. With the ongoing advancements in pre-trained large language models, it is possible that our method could be applied to a wider range of PLM types. Furthermore, we have only considered three commonly used perturbation types in this study, future studies could investigate a broader range of perturbations and how they interact with each other in determining the constituents. These limitations provide an opportunity to further improve the method and its applicability in the field.

\section*{Acknowledgements}

We would like to thank the anonymous reviewers, our meta-reviewer, and senior area chairs for their constructive comments. This research/project is supported by the Ministry of Education, Singapore, under its Tier 3 Programme (The Award No.: MOET320200004), and the National Research Foundation Singapore and DSO National Laboratories under the AI Singapore Program (AISG Award No: AISG2-RP-2020-016).

\bibliography{custom}
\bibliographystyle{acl_natbib}

\appendix
\section{Appendix}

\subsection{Dataset Statistics}
\label{appendix:dataset_stats}
Table \ref{tab:dataset_statistics} presents the statistical analysis of the English PTB and the multilingual SPMRL datasets. The table includes the number of sentences, the average sentence length, and the maximum sentence length from the development and test sets for each language. The data presented serves as an overview of the characteristics of the two datasets.

\begin{table*}[t]
    \begin{center}
        \centering
        \resizebox{1.0\linewidth}{!}{
        \small
        \begin{tabular}{lrrrrrrrrr}
            \toprule
                \textbf{Stats}  & \textbf{English} & \textbf{Basque}    & \textbf{French}    & \textbf{German}    & \textbf{Hebrew}    & \textbf{Hungarian} & \textbf{Korean}    & \textbf{Polish}    & \textbf{Swedish}    \\
            \midrule
                Size (development)                & 1,700  & 948   & 1,235  & 5,000  & 500   & 1,051  & 2,066  & 821   & 494 \\
                Avg. Length (development)         & 20    & 12   & 27    & 13    & 20    & 25    & 11    & 9     & 17 \\
                Max. Length (development)         & 98    & 37   & 98    & 60    & 87    & 76    & 26    & 26    & 101 \\
                Size (Test)                      & 2,416  & 946   & 2,540  & 4,999  & 716   & 1,009  & 2,287  & 822   & 666 \\
                Avg. Length (Test)               & 20   & 10    & 26    & 16    & 21    & 17    & 11    & 8     & 14 \\
                Max. Length (Test)               & 58   &31     & 119   & 115   & 70    & 56    & 29    &32     & 63 \\
            \bottomrule
        \end{tabular}
        }
        \caption{
            \label{tab:dataset_statistics}
            Dataset statistics of English PTB and the 8 languages from SPMRL. 
        }
        
    \end{center}
\end{table*}

\subsection{Performance Comparison by Layer}
\label{appendix:performance_layer}
\begin{figure}[t]
  \centering
  \begin{minipage}[c]{1.0\linewidth}
    \centering
    \begin{subfigure}{0.48\textwidth}
      \centering
      \includegraphics[width=\textwidth]{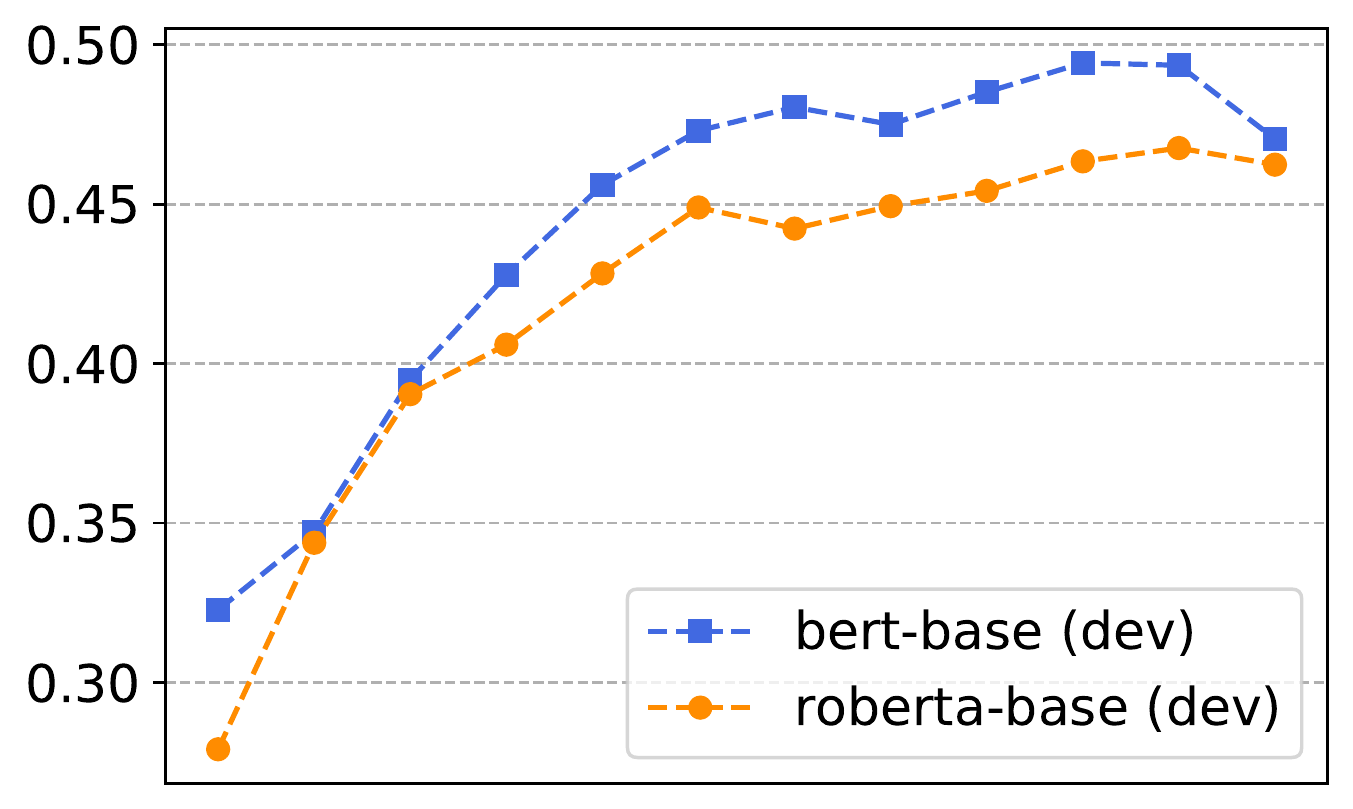}
      \captionsetup{font={scriptsize}}
      \caption{\label{fig:base_f1_layers} Base models with 12 layers }
    \end{subfigure}
    \begin{subfigure}{0.48\textwidth}
      \centering
      \includegraphics[width=\textwidth]{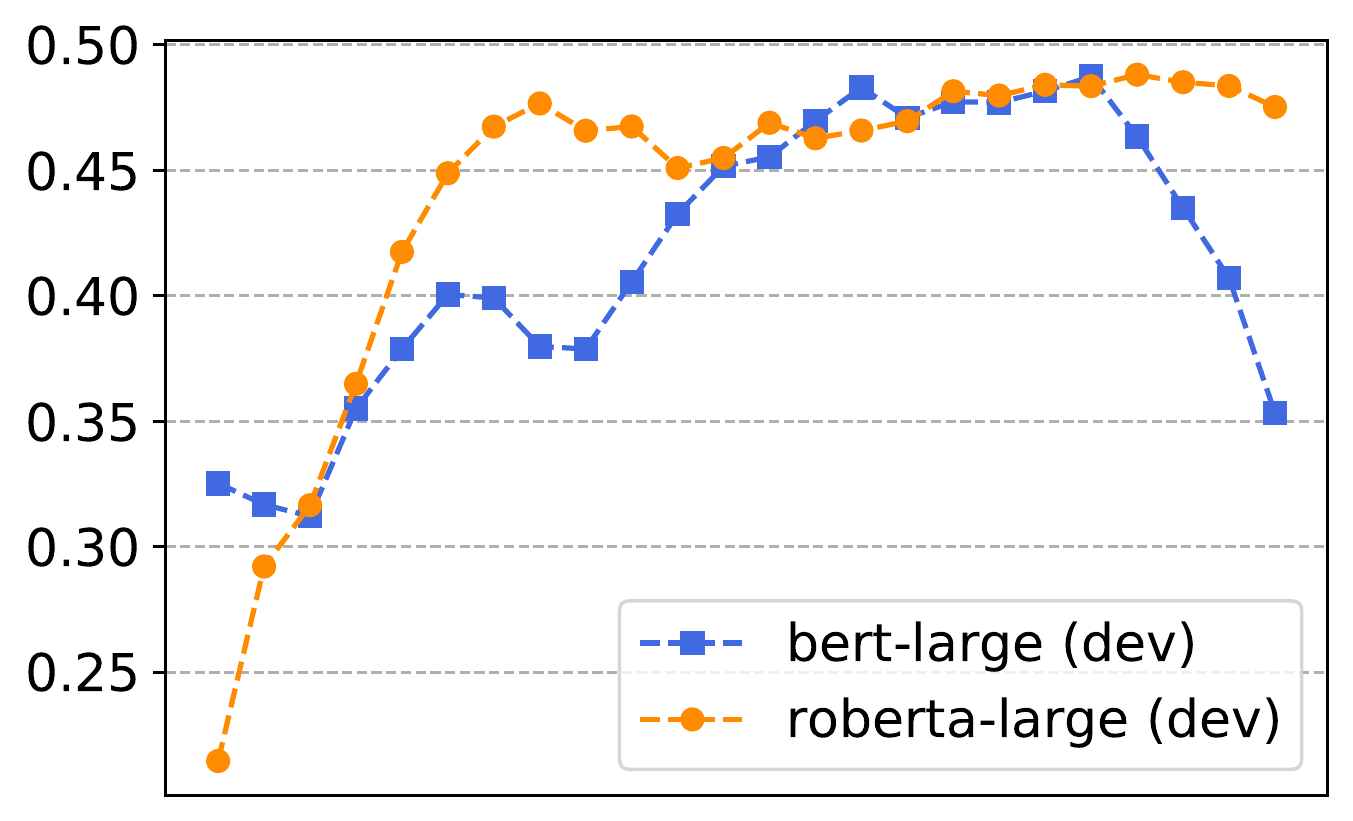}
      \captionsetup{font={scriptsize}}
      \caption{\label{fig:large_f1_layers} Large models with 24 layers. }
    \end{subfigure}
    \caption{The layer-wise sentence-level F1 scores on the PTB development set.}
    \label{fig:f1_layers_ptb_dev}
  \end{minipage}\hfill
\end{figure}

\begin{figure}[t]
    \centering
    \includegraphics[width=1.0\linewidth]{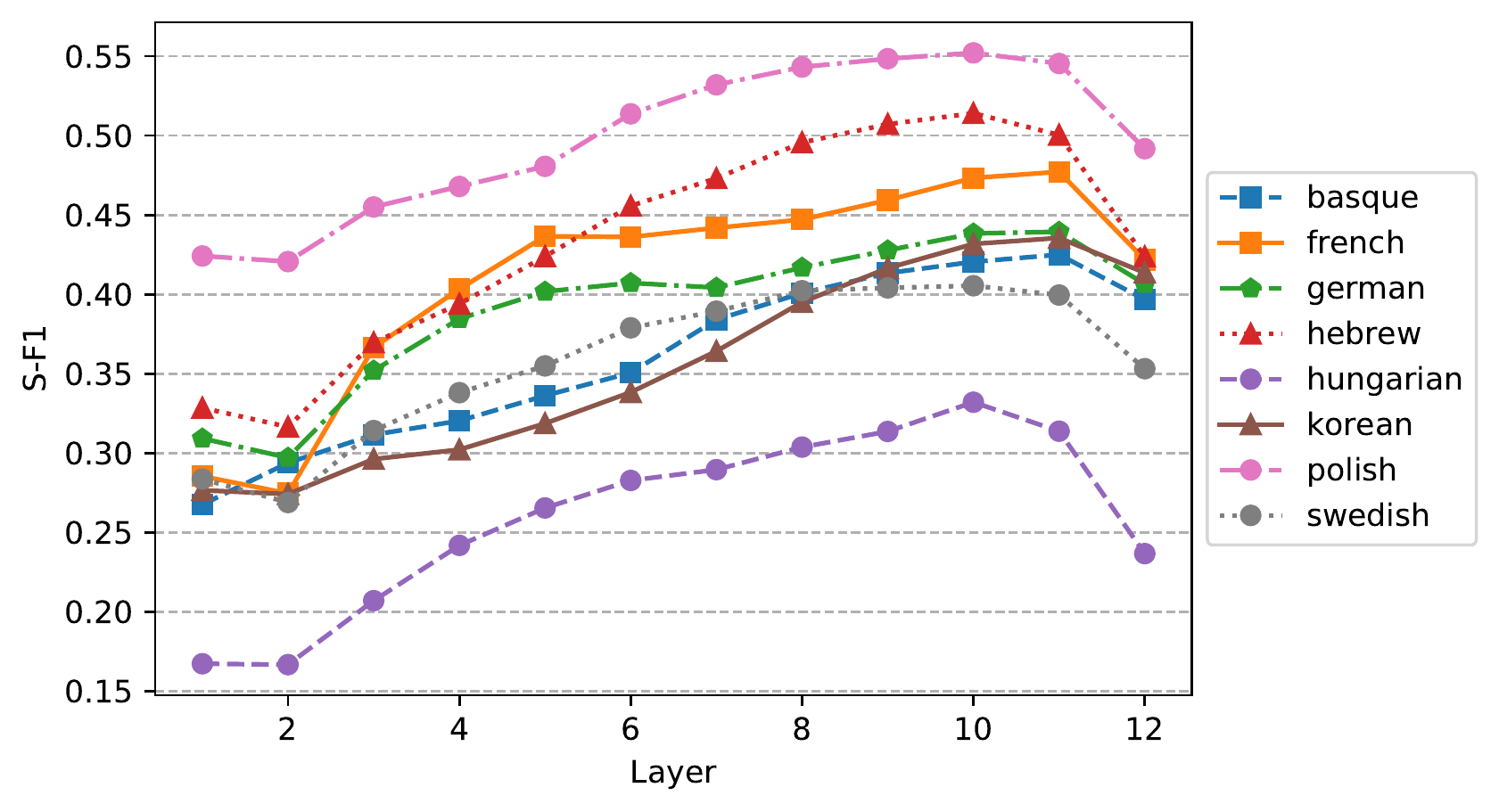}
    \caption{The layer-wise sentence-level F1 scores on the SPMRL development set. }
    \label{fig:f1_layers_spmrl_dev}
\end{figure}

This section presents the layer-wise performance of the English PTB and the multilingual SPMRL datasets on the development set. Specifically, for the English PTB, we evaluate the performance of BERT-base, BERT-large, RoBERTa-base, and RoBERTa-large models. For the multilingual SPMRL dataset, we use the M-BERT model. The results of the PTB development set are depicted in Figure \ref{fig:f1_layers_ptb_dev}, and the layer-wise comparison of the SPMRL development set is illustrated in Figure \ref{fig:f1_layers_spmrl_dev}. 

Our analysis shows that the pattern of performance for different languages is relatively consistent, with our method achieving the best results when using the later layers of the masked language models (LMs), although not necessarily the last layer. 
The performance over sentence lengths tends to increase until the last few layers. In our approach, the representation from LMs is mainly used for the semantic information it contains, as we focus on the change of contextual meaning due to perturbations on context. Thus, it is likely that the later layers contain richer semantic information.

On the development set of the SPMRL dataset, the best-performing layer for Basque, French, German, and Korean is layer 11, while the best-performing layer for Hebrew, Hungarian, Polish, and Swedish is layer 10. This suggests that there are specific layers, typically the later few layers in the LMs, which are more sensitive to linguistic perturbations and can reflect the contextual meaning of words. This highlights the importance of considering layer-wise representations when analyzing the performance of PLMs and the effect of context on their output when directly using the PLMs without finetuning.

\subsection{Impact of Perturbation Types}
\label{appendix:impact_perturbation_types}
In this section, we investigate the effect of various perturbation types on the development set of the SPMRL dataset. Table \ref{tab:ablation_perturb_spmrl} presents the sentence-level F1 scores for each language when one perturbation type is removed at a time. Overall, the best performance is obtained when three perturbations are applied. However, for Basque, German, and Polish, the best performance is achieved when two perturbations are combined. This suggests that certain languages may be more sensitive to certain types of perturbation. For example, removing the movement perturbation results in a 1.8 point increase in parsing performance in German. This can be attributed to the fact that word order is crucial in German in order to convey the correct meaning of a sentence, unlike in English where elements can be moved around without changing the overall meaning. Therefore, the movement perturbation may not be as effective in identifying constituents in German as it is in languages with more flexible word order, like English. Despite this, our method with all perturbation types does not significantly degrade the performance. For Basque and Polish, the performance with the combination of three perturbations is comparable to that of the best-performing combination. This demonstrates the robustness of our method across different languages and that all three perturbations generally improve parsing results.

\begin{table}[t!]
    \centering
    \small
    \begin{tabular}{lcccc}
    \toprule
        \textbf{Model} & \textbf{w/o sub} & \textbf{w/o dc} & \textbf{w/o move} & \textbf{all}   \\
    \midrule
        Basque      & 42.3  & \textbf{42.6}  & 31.8  & 42.5 \\
        French      & 46.9  & 43.2  & 41.8  & \textbf{47.7} \\
        German      & 42.2  & 41.0  & \textbf{45.7}  & 43.9 \\
        Hebrew      & 50.1  & 50.0  & 42.8  & \textbf{51.4} \\
        Hungarian   & 31.6  & 31.3  & 29.5  & \textbf{33.2} \\
        Korean      & 41.5  & 40.9  & 41.2  & \textbf{43.2} \\
        Polish      & \textbf{55.3}  & 51.8  & 47.4  & 55.2 \\
        Swedish     & 39.6  & 38.3  & 36.2  & \textbf{40.5} \\
        Avg.        & 43.7  & 42.4  & 39.6  & \textbf{44.7} \\
    \bottomrule
    \end{tabular}
    \caption{
        \label{tab:ablation_perturb_spmrl}
        Parsing performance with perturbation combinations on the SPMRL development set. 
    }
\end{table}

\begin{table*}[]
    \centering
    \small
    \begin{tabular}{llccccccc}
    \toprule
        \textbf{Method} & \textbf{Model} & \textbf{Sent F1} & \textbf{SBAR} & \textbf{NP} & \textbf{VP} & \textbf{PP} & \textbf{ADJP} & \textbf{ADVP} \\
    \midrule
    \multirow{4}{*}{Substitution} &
        BERT-base         & 25.4      & 36.3      & 28.7      & 18.6      & 34.8      & 30.3      & 43.3 \\
        & BERT-large      & 31.0      & 46.9      & 32.7      & 32.8      & 48.9      & 29.0      & 57.3 \\
        & RoBERTa-base    & 29.8      & 42.4      & 28.5      & 29.7      & 42.7      & 29.5      & 55.9 \\
        & RoBERTa-large   & 27.7      & 41.5      & 26.3      & 25.9      & 39.1      & 24.4      & 46.2 \\
    \midrule
    \multirow{4}{*}{Decontextualization} &   
        BERT-base       & 37.6      & 33.2      & 49.9      & 28.8      & 50.3      & 28.0      & 35.7 \\
        & BERT-large     & 39.7      & 29.6      & 54.6      & 26.5      & 55.1      & 33.6      & 45.6 \\
        & RoBERTa-base   & 33.8      & 26.8      & 44.0      & 25.8      & 37.9      & 26.6      & 31.8 \\
        & RoBERTa-large  & 32.5      & 28.8      & 42.4      & 26.7      & 36.7      & 30.9      & 24.8 \\
    \midrule
    \multirow{4}{*}{Movement} &  
        BERT-base       & 40.6      & 44.8      & 46.5      & 38.3      & 56.6      & 30.3      & 54.2 \\
        & BERT-large     & 39.1      & 44.5      & 44.3      & 39.1      & 51.1      & 27.3      & 51.7 \\
        & RoBERTa-base   & 43.5      & 49.4      & 51.7      & 39.4      & 59.3      & 42.8      & 55.6 \\
        & RoBERTa-large  & 45.9      & 52.5      & 54.6      & 40.5      & 66.9      & 40.9      & 61.5 \\      
    \bottomrule
    \end{tabular}
    \caption{Sentence-level F1 and label recall on the PTB test set for each individual perturbation.}
    \label{tab:perturbation-tag-recall}
\end{table*}

\begin{figure}[t!]
  \centering
  \begin{minipage}[c]{1.0\linewidth}
    \centering
    \begin{subfigure}{0.48\textwidth}
      \centering
      \includegraphics[width=\textwidth]{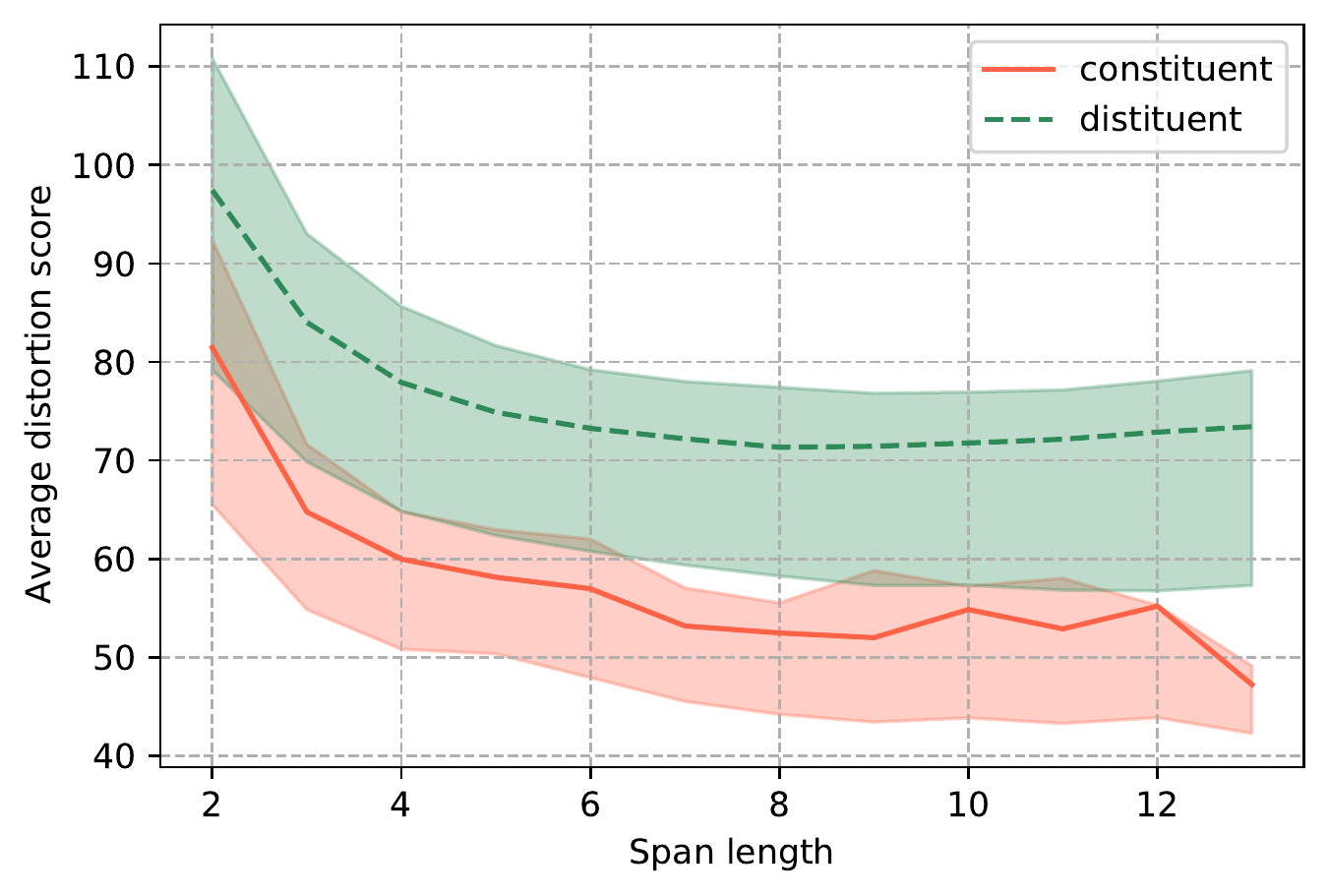}
      \captionsetup{font={scriptsize}}
      \caption{\label{fig:base_f1_layers} Basque. }
    \end{subfigure}
    \begin{subfigure}{0.48\textwidth}
      \centering
      \includegraphics[width=\textwidth]{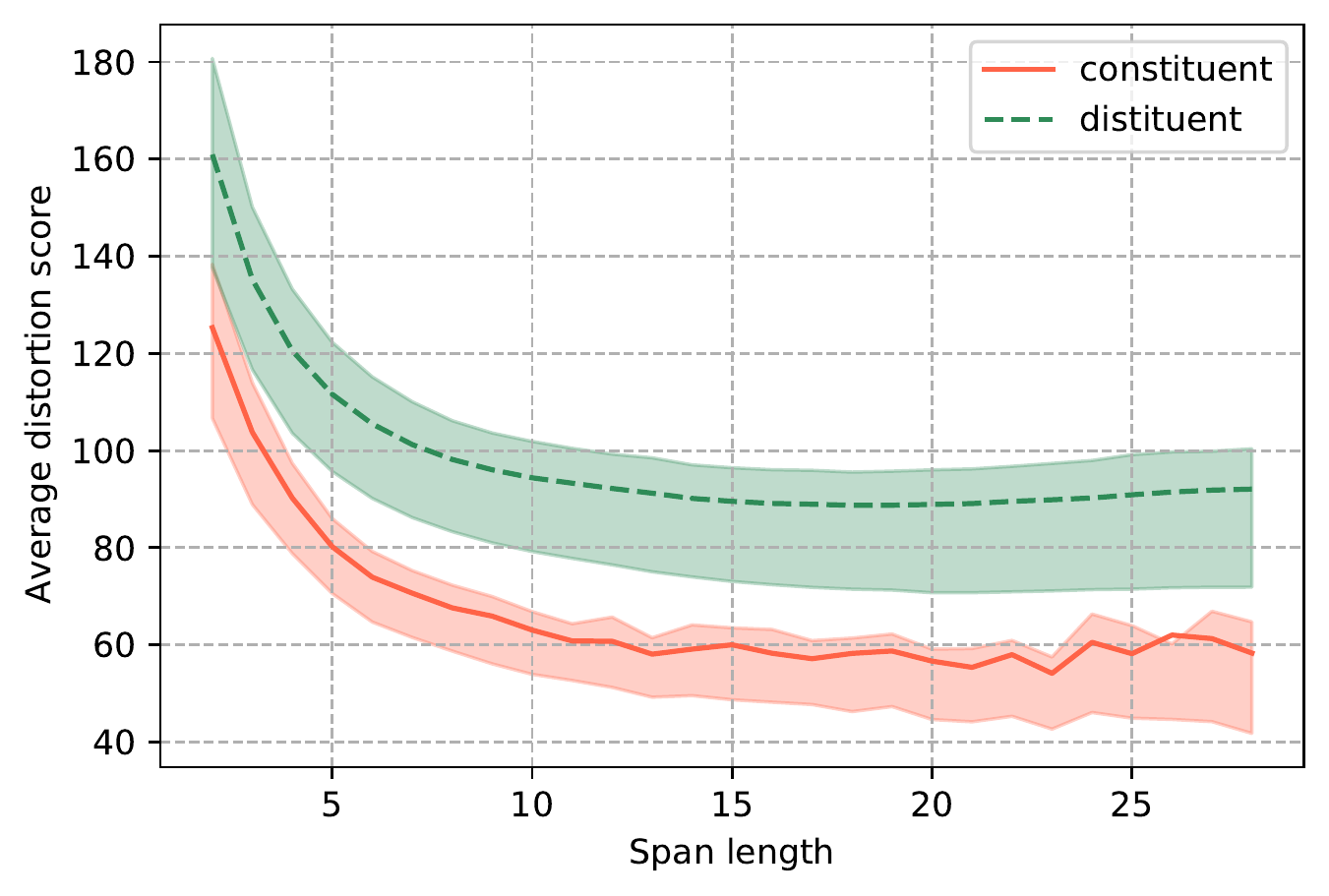}
      \captionsetup{font={scriptsize}}
      \caption{\label{fig:large_f1_layers} French. }
    \end{subfigure}
    \vskip\baselineskip
    \begin{subfigure}{0.48\textwidth}
      \centering
      \includegraphics[width=\textwidth]{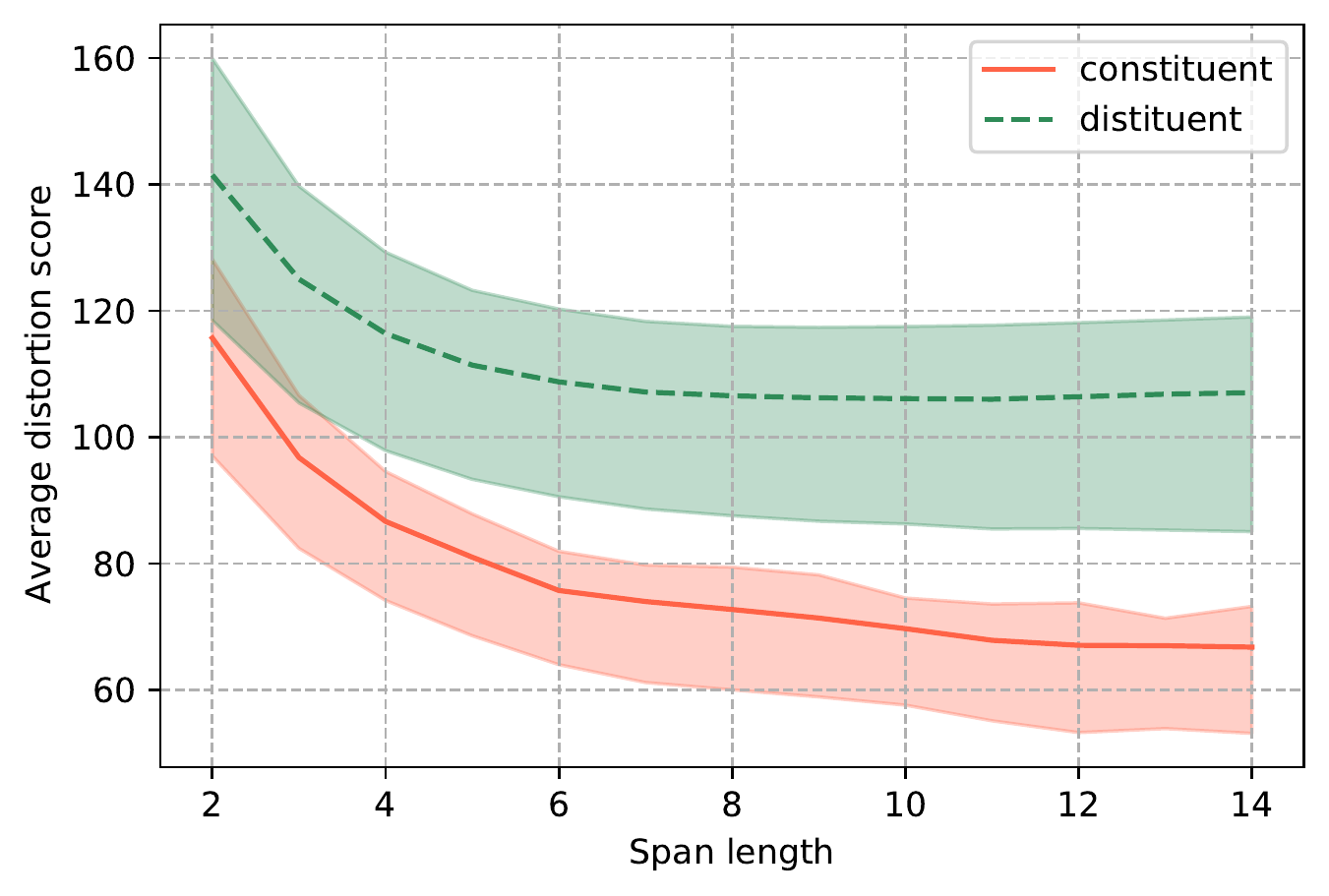}
      \captionsetup{font={scriptsize}}
      \caption{\label{fig:base_f1_layers} German.  }
    \end{subfigure}
    \begin{subfigure}{0.48\textwidth}
      \centering
      \includegraphics[width=\textwidth]{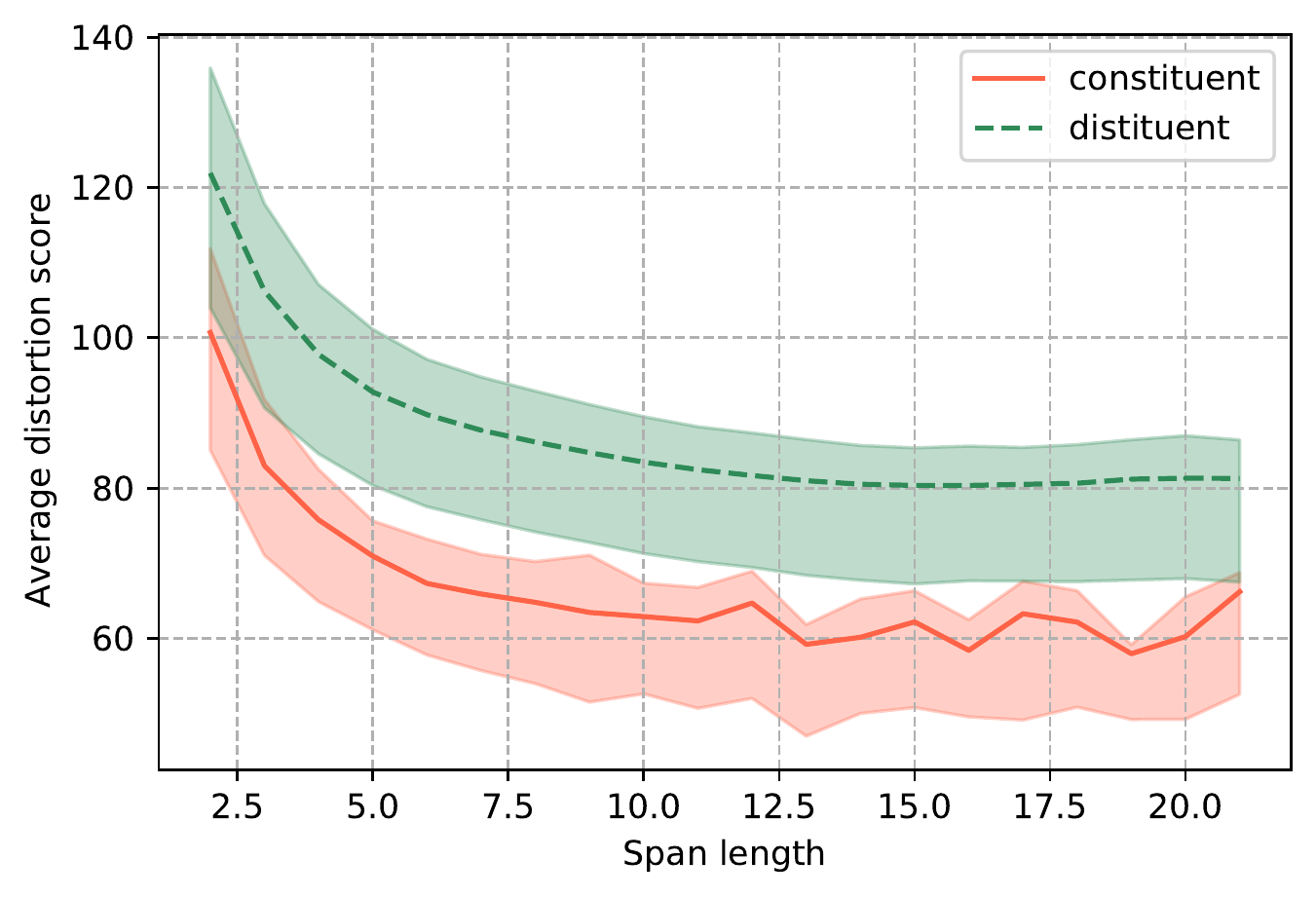}
      \captionsetup{font={scriptsize}}
      \caption{\label{fig:large_f1_layers} Hebrew.  }
    \end{subfigure}
    \vskip\baselineskip
    \begin{subfigure}{0.48\textwidth}
      \centering
      \includegraphics[width=\textwidth]{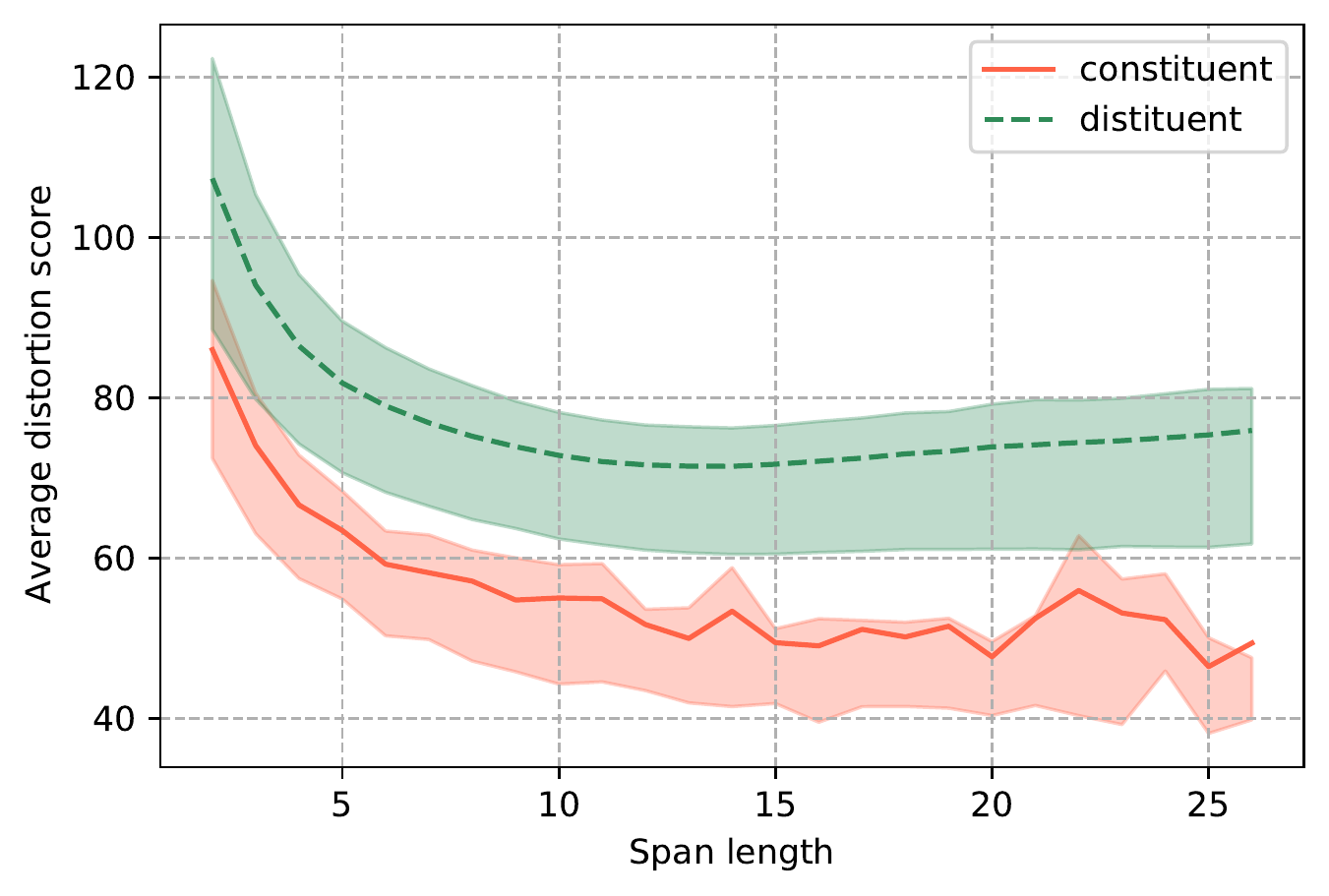}
      \captionsetup{font={scriptsize}}
      \caption{\label{fig:base_f1_layers} Hungarian.  }
    \end{subfigure}
    \begin{subfigure}{0.48\textwidth}
      \centering
      \includegraphics[width=\textwidth]{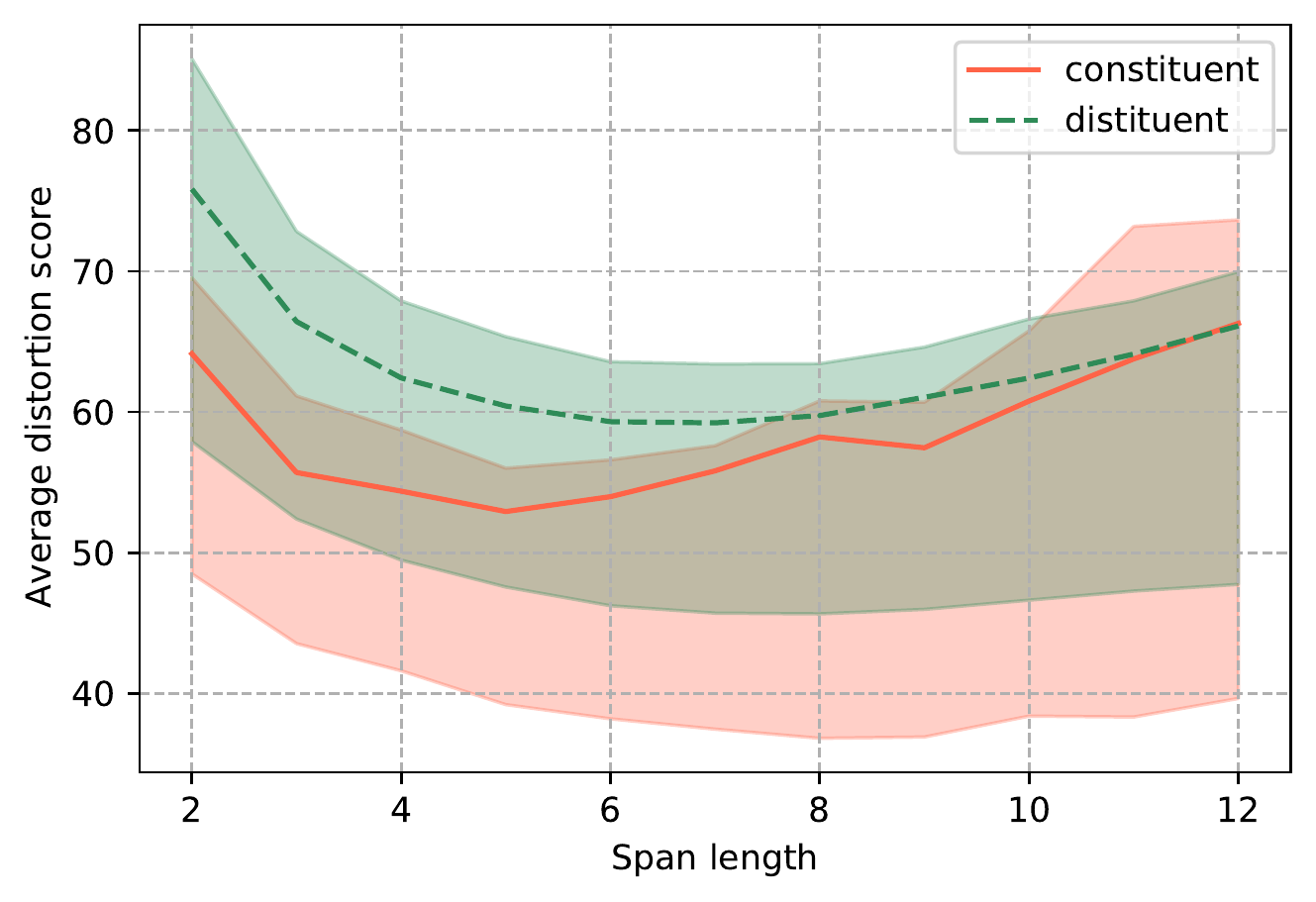}
      \captionsetup{font={scriptsize}}
      \caption{\label{fig:large_f1_layers} Korean.  }
    \end{subfigure}
    \vskip\baselineskip
    \begin{subfigure}{0.48\textwidth}
      \centering
      \includegraphics[width=\textwidth]{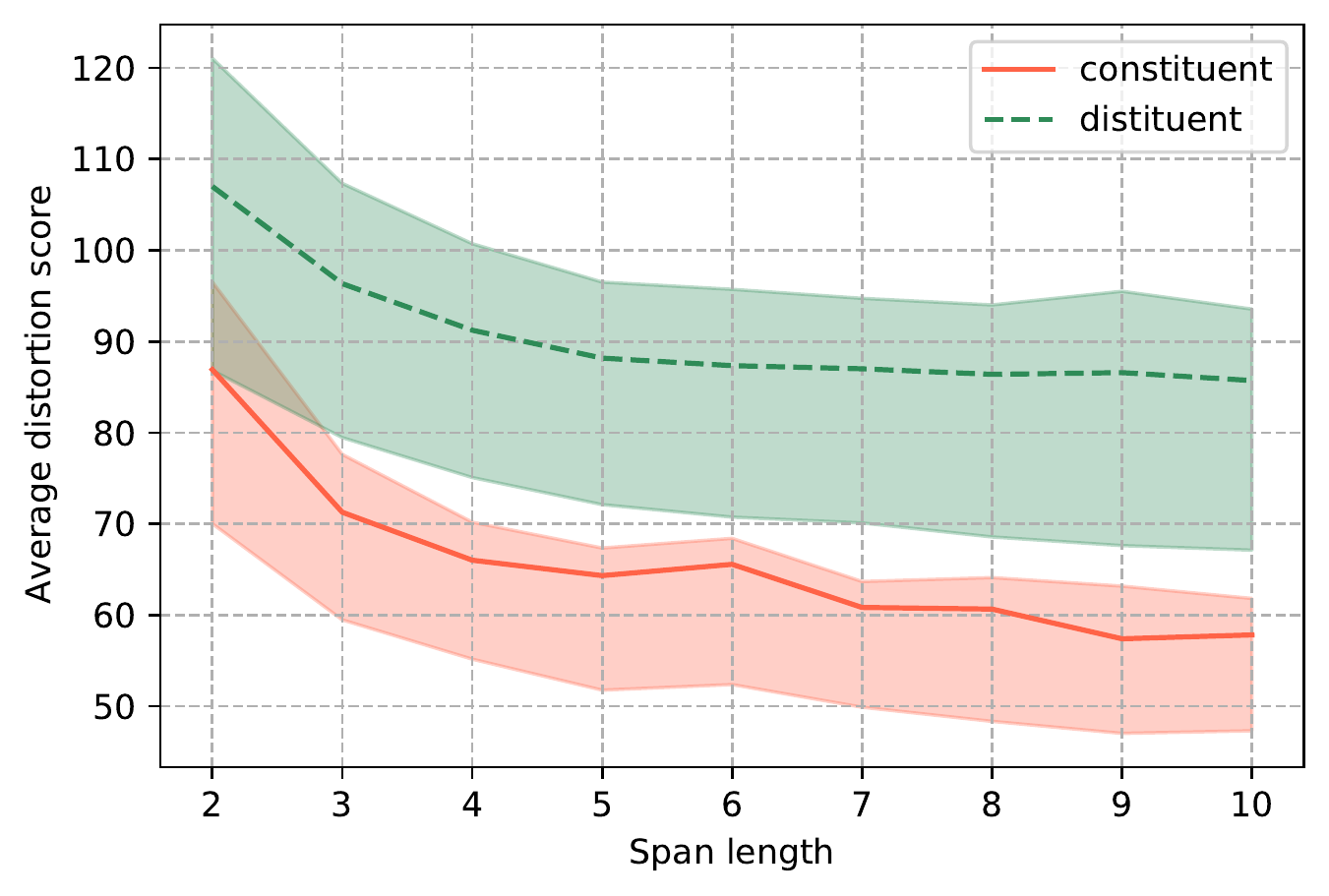}
      \captionsetup{font={scriptsize}}
      \caption{\label{fig:base_f1_layers} Polish.  }
    \end{subfigure}
    \begin{subfigure}{0.48\textwidth}
      \centering
      \includegraphics[width=\textwidth]{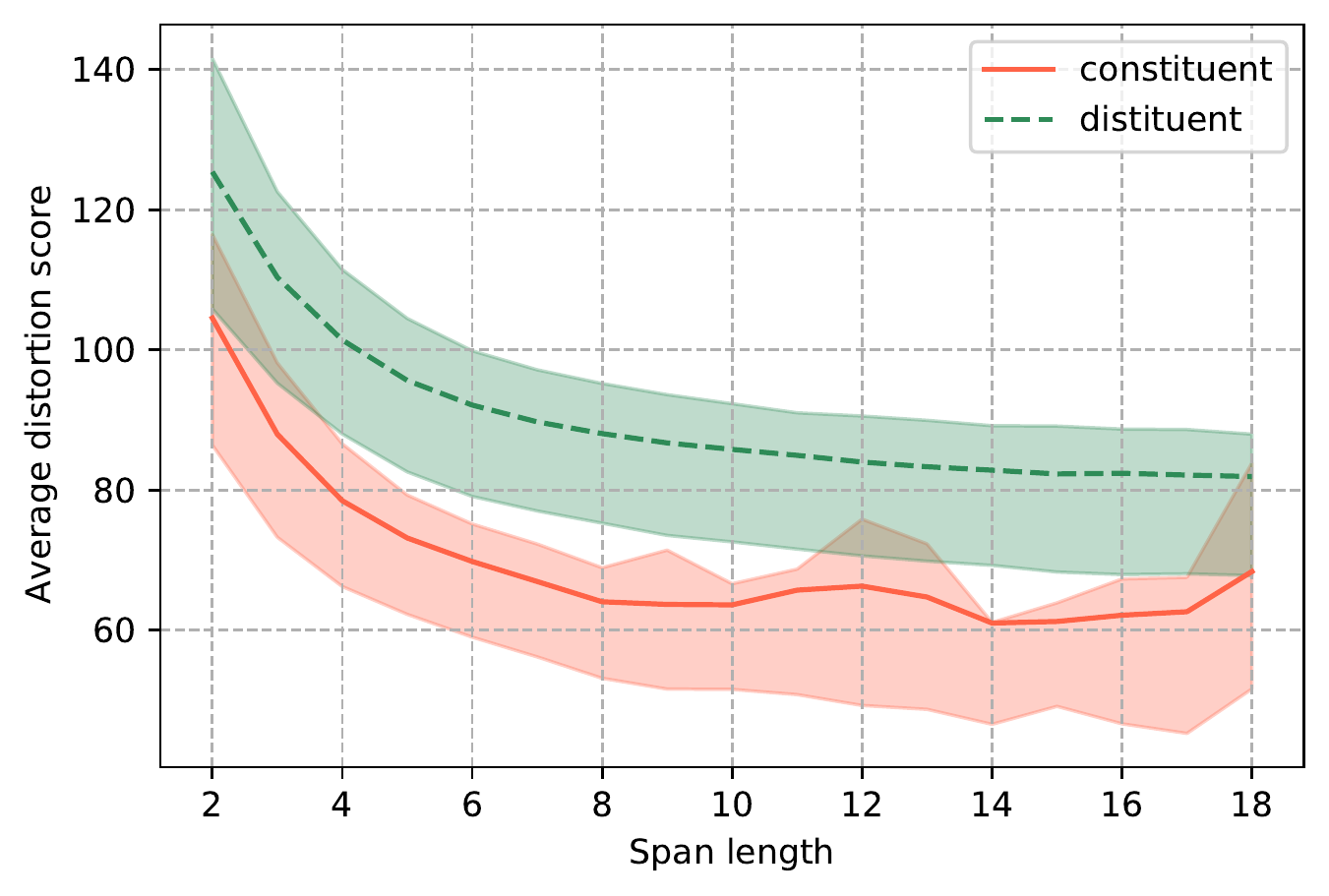}
      \captionsetup{font={scriptsize}}
      \caption{\label{fig:large_f1_layers} Swedish.  }
    \end{subfigure}
    \caption{The distortion scores of constituents and distituents for 8 languages in the SPMRL dataset.}
    
    \label{fig:distortion_spmrl}
  \end{minipage}\hfill
\end{figure}
\label{appendix:distortion_score_normalization}

We present the full results of label recall of 6 main constituency types when one perturbation is applied at a time on the English PTB test set, using BERT-base, BERT-large, RoBERTa-base, and RoBERTa-large in Table \ref{tab:perturbation-tag-recall}. The table demonstrates that movement perturbation is generally effective in capturing all 6 types of constituents compared to other perturbations. We note that each constituent type can be effectively captured by at least one perturbation and each perturbation targets different constituency types. The results for different models are consistent with some exceptions for certain constituency types. Further exploration of the differences in representation that lead to performance difference with respect to constituent types is left as future work. It is worth noting that our method with a single movement perturbation achieves comparable results to previous state-of-the-art methods. This highlights the effectiveness of the movement perturbation in detecting constituents. Similar findings can be observed in Table \ref{tab:ablation_perturb_spmrl} where the performance is the most affected without the movement perturbation.

\subsection{Distortion Score Reveals Constituency}

We analyze the correlation between the distortion score and constituency on the SPMRL development set. Figure \ref{fig:distortion_spmrl} illustrates the distortion scores for constituents and distituents of varying span lengths for each language. To ensure a sufficient number of constituents is considered in each length, we restrict our analysis to spans whose lengths are less than or equal to the average length of all sentences in the development set. The results show that the distortion scores for constituents are typically lower than those for distituents, providing evidence for our hypothesis that distortion scores calculated using our method can reveal the likelihood of a span being a constituent. This further suggests our method is robust across languages.

\subsection{Additional Implementation Details}
\label{sec:addtional_implementation_details}
We use the pre-trained LMs from a PyTorch codebase\footnote{{\url{https://huggingface.co/transformers/v3.3.1/pretrained_models.html}}}.

In instances where a word was split into word pieces, the representation of the word was obtained by averaging the representations of the word pieces.

We implement our method with PyTorch using Quadra RTX 8000 GPU. The estimated running time to parse the development set of English PTB with BERT-base is 1 hour.

\end{document}